\documentclass[final,1p,times]{elsarticle}
\usepackage{algorithm} 
\usepackage{algpseudocode}
\usepackage{enumitem}
\usepackage{amssymb}
\usepackage{amsmath}
\usepackage{subcaption}
\usepackage{etoolbox}
\usepackage{stfloats}
\usepackage{hyperref}
\usepackage{booktabs}
\usepackage{multicol,multirow}
\usepackage{colortbl}
\usepackage{xcolor}
\definecolor{TBLHeader}{gray}{0.85}
\definecolor{TBLRow}{gray}{0.95}
\usepackage{pifont}
\newcommand{\cmark}{\ding{51}}%
\newcommand{\xmark}{\ding{55}}%
\usepackage[switch,pagewise]{lineno}

\usepackage{lipsum}
\makeatletter
\def\ps@pprintTitle{
     \let\@oddhead\@empty
     \let\@evenhead\@empty
     \def\@oddfoot{\footnotesize\itshape
      Preprint accepted for publication in the \ifx\@journal\@empty Elsevier
      \else\@journal\fi\hfill October 2022} 
     \let\@evenfoot\@oddfoot}
\makeatother
  
\journal{Engineering Applications of Artificial Intelligence}

\begin{document}

\begin{frontmatter}

\title{IRMAC: Interpretable Refined Motifs in Binary Classification for Smart Grid Applications}

\author[UoA]{Rui Yuan}
\cortext[mycorrespondingauthor]{Corresponding author}
\ead{r.yuan@adelaide.edu.au}
\author[UoA]{S. Ali Pourmousavi}
\author[UoA]{Wen L. Soong}
\author[UoAm]{Giang Nguyen}
\author[watts]{Jon A. R. Liisberg}
\address[UoA]{The University of Adelaide, School of Electrical and Electronic Engineering, Adelaide, South Australia, Australia.}
\address[UoAm]{The University of Adelaide, School of Mathematical Sciences, Adelaide, South Australia, Australia.}
\address[watts]{Watts, Copenhagen, Denmark.}




\begin{abstract}
Modern power systems are experiencing the challenge of high uncertainty with the increasing penetration of renewable energy resources and the electrification of heating systems. In this paradigm shift, understanding electricity users' demand is of utmost value to the retailers, aggregators, and policymakers. However, behind-the-meter (BTM) equipment and appliances at the household level are unknown to the other stakeholders mainly due to privacy concerns and tight regulations. In this paper, we seek to identify residential consumers based on their BTM equipment, mainly rooftop photovoltaic (PV) systems and electric heating, using imported/purchased energy data from utility meters. To solve this problem with an interpretable, fast, secure, and maintainable solution, we propose an integrated method called Interpretable Refined Motifs And binary Classification (IRMAC). The proposed method comprises a novel shape-based pattern extraction technique, called Refined Motif (RM) discovery, and a single-neuron classifier. The first part extracts a sub-pattern from the long time series considering the frequency of occurrences, average dissimilarity, and time dynamics while emphasising specific times with annotated distances. The second part identifies users' types with linear complexity while preserving the transparency of the algorithms. With the real data from Australia and Denmark, the proposed method is tested and verified in identifying PV owners and electrical heating system users. The performance of the IRMAC is studied and compared with various state-of-the-art methods. The proposed method reached an accuracy of 96\% in identifying rooftop PV users and 94.4\% in identifying electric heating users, which is comparable to the best solution based on deep learning, while the speed of the inference model is a thousand times faster. Last but not least, the proposed method is a transparent algorithm, which can tackle the concerns regarding the agnostic decision-making process when policies prohibit some machine learning methods.
\end{abstract}

\begin{keyword}
\texttt{Motif discovery, Data analytic, Renewable energy, Pattern recognition, Binary classification, Time series data mining}
\end{keyword}

\end{frontmatter}

\section{Introduction}\label{intro}
\subsection{Background}
Advances in distributed renewable energy generation and electrification led to a rapid growth of uncertainty in modern power system operation and planning. For example, consumers' net demand is heavily affected by rooftop PV generation uncertainty, which makes it cumbersome to estimate electricity demand. As a result, knowing residential consumers with rooftop PVs and electric heating systems at the feeder level can play an important role in infrastructure planning, day-to-day grid operation, network upgrade, and demand-side management. However, information on whether residential users own these types of equipment is unavailable for many reasons, e.g., policy and privacy concerns \cite{Kop2021}. In some countries, e.g., the USA \cite{Barbose2021}, there is a state-owned database to keep records of the home solar systems, which retailers can access. Also, in jurisdictions with Feed-in-Tariff mechanisms, users with rooftop PV systems are known to the retailer \cite{Australia2008}. However, such databases are either incomplete (e.g, in the case of the USA where more than 21\% of PV owners are not listed) or do not exist in some jurisdictions where users installed unqualifying generators that is defined as a photovoltaic system with capacity over 10kVA for a single phase connection or over 30kVA for a three phase connection \cite{Barbose2021,Australia2008}. Additionally, other entities like aggregators may not have access to the database of registered solar users owing by one retailer because of regulations \cite{AGLEnergyLimited2015}. On the other hand, such database does not exist for the users with electric heating systems. Furthermore, identifying users from static databases and records may not be helpful to all stakeholders. For instance, a demand response operator or an aggregator appreciates a dynamic users identification approach, where faulty, shaded, or under-performed PV systems are also identified. 

The most accessible and reliable data for all stakeholders to identify types of consumers is the users' purchased electricity data collected by their utility meters, known as imported electricity because of two reasons: 1) Some utilities in the world, e.g., Watts in Denmark, who is our industry partner, only have access to imported data from the grid for their consumers, and 2) Users with only conventional applications like electric heating systems do not have exported electricity data. However, identifying users types based on their hourly or half-hourly imported electricity from the grid is a non-trivial problem, partially because of the substantial amount of time series data from a large number of consumers. The complexity grows exponentially when the queried time range increases and more consumers are required to be analysed. In addition to the computational difficulty, electricity demand patterns tend to change over the year, influenced by weather conditions (e.g., sunny or cloudy days and daylight hours length) and consumers' behaviour \cite{MoradiSizkouhi2021}. Furthermore, there are patterns in the imported electricity data that do not contain the required features, e.g., nighttime hours for solar or summer hours for electric heating. Those irrelevant patterns deteriorate the performance of intuitive methods, e.g., load duration, counting zeros, and average profiles; thus, they are ineffective in solving this problem, as we will show in Section \ref{experiment}. Last but not least, deep learning methods and other over-parametrized machine learning (ML) models show improved generalisation on complex datasets \cite{Power2021,Balaji2021}. However, those methods are arguably inappropriate or unacceptable to be used in power systems as their decision-making processes are not fully transparent \cite{Kamath2021, Kop2021}. In 2021, European Commission presented the EU Artificial Intelligence Act, which states the transparency of ML methods needs to be carefully assessed and justified before using on critical infrastructures like power systems, as it could put the life and health of citizens at risk~\cite{EUact, Kop2021}. As a result, interpretability has become a mandatory consideration for applications of safety-critical systems. Consequently, a fast, accurate, scalable and interpretable solution is needed for this classification problem.
\subsection{Related works}
Classification problems with long time series have been discussed in the literature for decades. The traditional classification methodologies can be categorised into two groups: shape-based and structure-based \cite{Fang2018, Ding2008}. The former approaches rely on analysing similarities of patterns of the raw numeric time series, while the latter converts the raw data with Symbolic Aggregate Approximation (SAX) or the Discrete Fourier Transform (DFT) to create statistical models. Besides the shape-based and structure-based methods, some research studies classified electricity consumers by analysing their electricity consumption time series with ML techniques, e.g., \cite{Bidoki2010, Peng2016, Butunoi2017, Wu2020, Wus2021}. While ML techniques can generally handle extremely complex systems and infer from incomplete data, their application in power system operation as a safety-critical system casts doubt. ML-based models cannot be easily interpreted, their behaviour cannot be anticipated, they are vulnerable to false data injection attacks, and many of them neglect the domain knowledge and physical models \cite{Guidotti2018, goodman2016eu, Korkmaz2022}, except for some recent efforts to add physical models to neural networks \cite{Stiasny2021,Nellikkath2021,Elsheikh2019}.
Furthermore, the enormous number of parameters in most ML techniques, e.g., Deep Neural Network (DNN), make them suffer from the curse of dimensionality and long training time \cite{10.1007/11494669_93, Wu2022}. On the contrary, the shape-based methods project the data into a much lower relevant representational space to eliminate the curse of dimensionality \cite{10.1007/11494669_93}. The authors in \cite{Funde2019} discussed how shape-based methods can be used in energy consumption data analysis for reliable and interpretable applications. Shape-based approaches are believed to be more accurate and interpretable but are computationally expensive \cite{Fang2018, Mills2006}. One solution to balance the advantages and drawbacks of the above techniques is called \emph{motif extraction}, which can compress the time series data while preserving the shape information for classification purposes in an interpretable manner.

Motifs, defined as approximately repeated sub-patterns in a long time series, were first proposed in 2003 \cite{chiu2003probabilistic}. Since then, motifs have been used as representative patterns for long time series data in various data mining applications, e.g., classification, clustering, and rule discovery \cite{Gonzalez-Alvarez2013, Fu2011}. However, efficient ways to extract motifs were needed as the brute-force solution was computationally untenable \cite{Linardi2020}. In 2016, an all-pair-similarity search technique, called \emph{Matrix Profile} (MP), was proposed and then widely used as it could significantly decrease the spatial and temporal complexity of the motif discovery problem \cite{Yeh2017}. The core idea of MP is to record the most similar sub-patterns pair with the smallest z-normalised Euclidean distance (ED) and exploit the overlap between subsequent patterns using Fast Fourier Transform (FFT) \cite{Yeh2017}. MP then evolved into a Nearest-Neighbour-based approach and became the most dominant motif discovery approach in the literature, while the definition of motif was changed into the closest pair of subsequences \cite{Linardi2020,Yeh2017,Akbarinia2019,Zimmerman2019}. Another critical improvement occurred in 2018, when the Scalable Time series Ordered-search Matrix Profile (STOMP) algorithm was proposed, which further significantly decreased the temporal complexity of motif discovery by enabling parallel computing and GPU acceleration \cite{Zhu2018}. Since then, motif-based classification techniques have been widely used in finance, bioinformatics, and economics, due to their robustness, low computational power requirement and interpretability. Notably, researchers aimed at proposing a general solution for users with no prior domain knowledge in \cite{Imani2018, Yeh2017}, while this general solution and classic motifs are arguably limited in real-world experiments when domain-dependent knowledge is expected to adjust the mismatch between extracted motifs and user's expectations \cite{Dau2017}.

Compared to the research fields above, few studies have investigated the application of motif-based classification methods in power systems \cite{Funde2019, Funde2018, Ling2020}. In these studies, classic motif-based discovery methods were applied to smart meter energy consumption data to identify regular behaviours, rules extraction, and solar PV panel installation identification. All three papers used SAX to represent the time series data as a discrete code sequence, from which the repeated code segments were extracted as motifs. However, as discussed above, similar to motifs, SAX is a representation technique for dimensionality reduction \cite{Cassisi2012, Wang2016, Yu2019, Fang2018}. The issue with this approach is the loss of shape information critical in binary classification problems such as the one in this paper. Furthermore, SAX is vulnerable to identifying small amplitude changes on sub-patterns and missing important information in a given segment \cite{Yu2019}. 

\subsection{Objectives and contributions}
As motifs-based applications have become feasible only in the last couple of years, there are some knowledge gaps in state-of-the-art motif discovery methods. First, current motif discovery methods always extract the most significant features and hence are unsuitable as a general tool in different applications \cite{Akbarinia2019, Linardi2020, Zimmerman2019}. Second, extracting exact repeated patterns with domain knowledge still remains an unsolved problem. Last but not least, there is a lack of discussion on how to effectively use the extracted motifs and preserve the high flexibility for classification purposes in the literature.

To fill the knowledge gap, we propose a highly flexible motif discovery method, which enables users to bring in domain knowledge to identify the most occurring sub-patterns. To the best of our knowledge, this is the first motif discovery method that can extract motifs containing the requested features that might not be the most significant to the shape whilst considering the time dynamics and preserving the interpretability. With the extracted motifs as inputs, we construct a linear single-neuron model to classify users by their motifs. This methodology is later used in a comprehensive simulation study to solve the rooftop PV owners' identification problem in this paper, with high performance on accuracy and speed. To show the robustness of the proposed method, a second problem is solved by identifying electric heater users in a real dataset. The methodological contribution of the paper includes a novel sub-pattern extraction technique and a hierarchical structure for residential PV users' identification. The former is called \emph{Refined Motif discovery} which detects the most repeated pattern by the proposed \emph{Similarity Profile} that computes similarity indices of each sub-pattern with pairwise annotated Dynamic Time Warping (DTW) distances. The latter is computing the Refined Motif discovery at the users' end in parallel while implementing the classification method in the cloud. With the help of smart meters, applying the proposed method in the modern power system has seven advantages: memory efficiency, low communication band-with requirement, high computation speed, cost efficiency, robustness to abnormal data, low privacy concerns, and easy maintenance, which are discussed in detail in Section \ref{method}. In summary, the significance of this paper is outlined as follows:
\begin{itemize}
    \item A novel motif discovery method to extract the most occurring sub-patterns considering domain knowledge, temporal dynamics, and the average similarity. The RM discovery process is application oriented, meaning it measures the sub-patterns similarity based on the application instead of extracting the most significant features.
    \item A systematic solution of finding RM at the users' end and sending the RMs to the cloud, where a linear classifier identifies the users' type. The process is transparent and has several advantages for utilities and end-users.
    \item A comprehensive performance comparison among the proposed method, domain knowledge-based methods, ML-based and other motif based methods considering accuracy, speed, and transparency. Two datasets for identifying rooftop PV and electric heating systems are tested and assessed.
\end{itemize}
This paper is organised as follows: Section \ref{method} explains the proposed methodology including motif discovery, classification, and implementation. The simulation studies are reported in Section \ref{experiment}, and the results are analysed in detail. The paper is concluded in Section \ref{future} and future works are outlined.

\section{The proposed methodology}
\label{method}
\begin{figure}[!ht]
\centering
\includegraphics[clip, trim=0cm 0cm 0cm 0cm, width=5in]{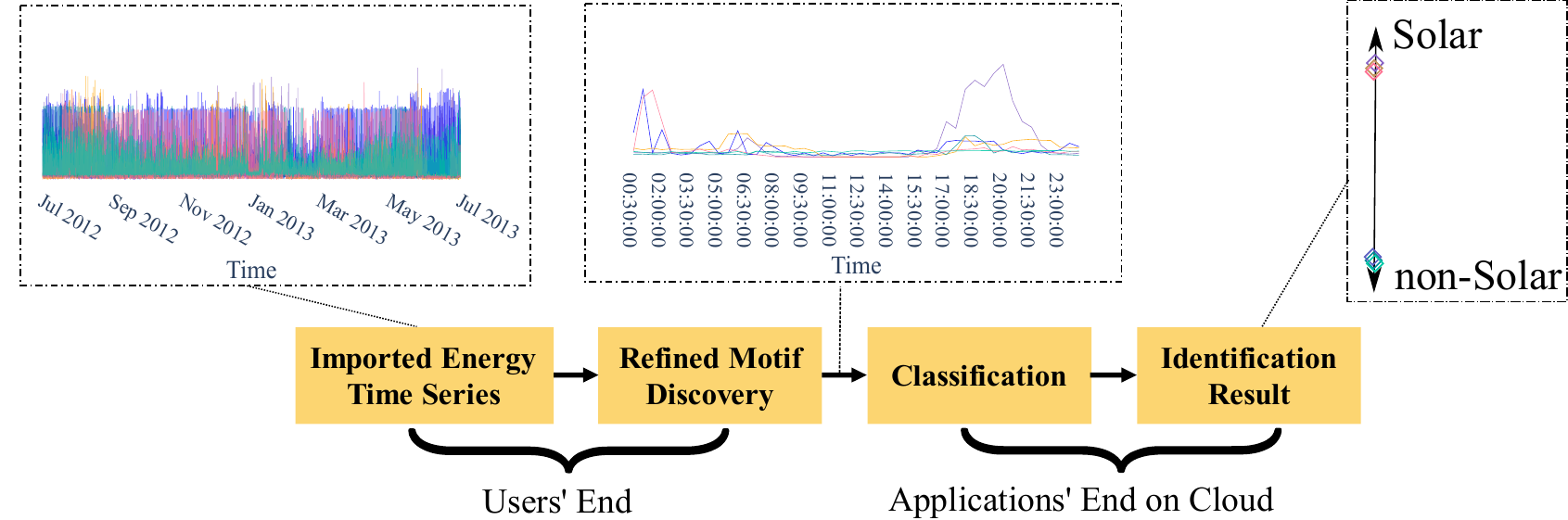}
\caption{A schematic showing the processing steps of the proposed classification method with 6 sample users}
\label{fig_block_overall}
\end{figure}

The processing steps of the proposed methodology is summarised in the schematic of Figure \ref{fig_block_overall}. 
First, an accurate and fast sub-pattern extracting method is developed to reduce the dimension of the electricity time series. In this method, a different similarity measure is proposed by combining power system domain knowledge with temporal dynamics of the time series to extract features. In addition, unlike the Nearest-Neighbour-based methods looking for the closest patterns as motifs, this technique discovers the most \emph{repeated} patterns during the time intervals of interest, which is called \emph{Refined Motif} (RM) hereafter. For instance, in the rooftop PV identification problem, each user's most repeated sub-pattern is extracted from one year of data considering daytime features, as in Figure \ref{fig_block_overall}. Second, with the features discovered by the RM method, a linear-complex classification model is proposed to identify users' types, which includes weight parameters and a threshold to provide classification results, e.g., the solar user or non-solar user, as shown in Figure \ref{fig_block_overall}. 
Section \ref{sec: RM-discovery} and Section \ref{ML} will introduce the two middle steps in detail. Section \ref{sec: systematic advantages} addresses a systematic solution of computing RM discovery at users' end and implement the classification process at applications' end. 
\subsection{Refined Motif (RM) Discovery Method}
\label{sec: RM-discovery}
One issue in the previous motif discovery methods is the lack of temporal dynamics representation.
Although DTW has long been proven to outperform the ED measure in terms of accuracy and detecting temporal dynamics, i.e., time shifting and time stretch \cite{Ding2008}, the ED was the preferred similarity measure in motif discovery research, e.g., \cite{Linardi2020, Yeh2017, Akbarinia2019, Zimmerman2019, Zhu2018}, due to lower complexity. The ED and DTW measures are given in Equation~\eqref{eq:Ed} and Equations~\eqref{eq:DTW}-\eqref{eq:DTW-2}, respectively, where two sub-patterns $\mathbf{X}$ and $\mathbf{Y}$ with the same length, $m$, are compared by ED and DTW: 
\begin{align}
    \text{ED}(X, Y)=\sqrt{\sum^m_{i=1}(X_i-Y_i)^2}\label{eq:Ed}
\end{align}
DTW finds the minimum warping path and its associated distance of the two sub-patterns with Equations~\eqref{eq:DTW}-\eqref{eq:DTW-2}:
\begin{equation}
    \text{DTW}(X, Y) = \sqrt{\Theta_{X_{m},Y_{m}}}\label{eq:DTW}
\end{equation}
\noindent where $X_m$ and $Y_m$ are the $m$\textsuperscript{th} point in $X$ and $Y$, $\Theta_{X_{m},Y_{m}}$ is the cumulative distance of $X$ and $Y$ from $1$ to $m$. The distance between $i$\textsuperscript{th} point in $X$ and $j$\textsuperscript{th} point in $Y$ can be calculated as
\begin{align}
    \Theta_{X_i,Y_j} = (X_i-Y_j)^2+\min\{\Theta_{X_{i-1},Y_{j-1}}, \Theta_{X_{i-1}, Y_j}, \Theta_{X_i,Y_{j-1}}\}\label{eq:DTW-2}
\end{align}

In this case, the complexity of ED is $O(m)$, while it is quadratic, $O(m^2)$, for DTW. Consequently, for a long time series with total length $n$ and sub-pattern length $m$ with one data point sliding step, the motif discovery complexity is $O(n^2m)$ and $O(n^2m^2)$ for ED and DTW, respectively. Furthermore, z-normalised ED between two time series sub-patterns can be calculated by their dot product, hereby using the Fast Fourier Transform (FFT) divide and conquer to reduce the complexity from $O(n^2m)$ to $O(n^2\log m)$ (best case $O(n^2)$), which is called the MASS algorithm \cite{Yeh2017}. Although some efforts have been done on computationally-improved DTW algorithms, e.g. Dynamic Timewarp Barycenter Averaging and Fast Dynamic Time Warping (FDTW) \cite{chan2018evaluation}, they sacrifice the accuracy for a higher computational speed, hence are less accurate than DTW and slower than ED. In addition, one research paper argued that the most cited FDTW method in the literature is slower than the original DTW \cite{Wu2021}. Some studies compute DTW-based motifs by omitting some computations with lower bounding and abandoning techniques, which compromise the information on similarities of every sub-patterns pairs \cite{Lagun2014,Wu2021,Alaee2020,FurtadoSilva2019}. 

In this paper, we use the DTW measure instead of the ED measure for two reasons. First, the nature of specific problems, such as PV owners' identification, necessitates more complex distance measures. For instance, seasonality will affect the PV generation magnitude and duration, such that two clear-sky days in summer and winter may be seen as two different patterns using simple distance measures due to the patterns' stretches. Second and most importantly, the computational speed of the proposed method is similar to ED-based motif discovery with the help of domain knowledge, while neither bounding nor abandoning techniques are applied as the average similarity of each sub-pattern is considered.
 

Because a PV system generates energy in a daily repetition, the sub-pattern size $m$ should equal the number of samples in the daytime. Also, PV systems generation profiles present a daily cycle. Therefore, we change the pattern search sliding window length from $1$ to $m$. The wider sliding window requires the pattern similarity measurement to be robust to pattern shifting due to the seasonal changes in the sunrise and sunset times. As a cumulative distance measure, DTW works perfectly under time shifting. At the same time, a wider sliding window means only ${n}/{m}$ steps are required instead of $n$ to assess the entire time series, which significantly reduces computational time to ${1}/{m^2}$ of the signal step sliding where the computation is pairwise. As a result, the time complexity of the DTW-based motif discovery scales down from $O(n^2m^2)$ to $O(n^2)$, similar to the MASS algorithm. More importantly, it makes every sub-pattern start and end at the same time, which ensures that the motifs extracted from different users are comparable. 

Another issue in the existing motif discovery methods is that one can search only for the most significant feature inside the sub-pattern. As a result, extracting motifs from one year of imported electricity data potentially leads to finding the most repeated consumption pattern that dominates the 24-hour daily cycle. This is because the PV generation is only available during daytime, the patterns of which may be undermined by the stronger 24-hour load patterns. 
We show in Section \ref{experiment} that conventional motif discovery methods cannot extract meaningful sub-patterns in such cases. In this regard, we propose an annotated DTW to measure the distance between sub-patterns by emphasising specific temporal patterns during the daylight hours, as \eqref{eq:rDTW}-\eqref{eq:rDTW-2}, where $X$ and $Y$ are two sub-patterns with length $m$, $\Theta'_{X_{i},Y_{j}}$ is the cumulative distance of point $i$ on $X$ and $j$ on $Y$, and $W$ is a proposed weight matrix for customising the extracted features by assigning different weight to temporal data:

\begin{equation}
    d(X,Y) = \text{DTW}(X, Y) = \sqrt{\Theta'_{X_{m},Y_{m}}} \label{eq:rDTW}
\end{equation}
\begin{equation}
    \Theta'_{X_{i},Y_{j}} = W_{i,j}(X_i-Y_j)^2+\min\{\Theta'_{X_{i-1},Y_{j-1}}, \Theta'_{X_{i-1},Y_j}, \Theta'_{X_i,Y_{j-1}}\} \label{eq:rDTW-2}
\end{equation}
\begin{align}
W_{i,j} = \max\left\{|\vec{w}_{i}|, |\vec{w}_{j}|\right\} \label{eq:mask1}\\
\vec{w}_{t} = 
  \begin{cases}
    1, \;\;t\in \text{daytime}\\
    0, \;\;t\in \text{nighttime}
    \end{cases}\label{eq:mask2}
\end{align}

To use annotated DTW for PV pattern identification, we need to emphasise the daytime patterns. As a result, the weighted matrix $W_{i,j}$ in Equation~\eqref{eq:rDTW-2} is set as binary in Equations~\eqref{eq:mask1} and \eqref{eq:mask2}, where $w_t$ is a binary mask vector with 1 in the daytime and 0 in the nighttime. This can further reduce our motif discovery complexity from $O(n^2)$ to $O(n^2{m'}/{m})$, where $m'$ is the length of daytime. For solar users' identification, we set 9am to 4pm as daytime considering the overlapped period under the four seasons. For electric heating users' identification, no mask was applied in the experiment due to the lack of comprehensive information, but theoretically, the weighted matrix can be set based on their heater usage probability distribution. This annotated DTW method is used to calculate the distance between two patterns of the day $X$ and day $Y$ in the given queries.

\begin{figure}[!ht]
\centering
\includegraphics[width=3.1in]{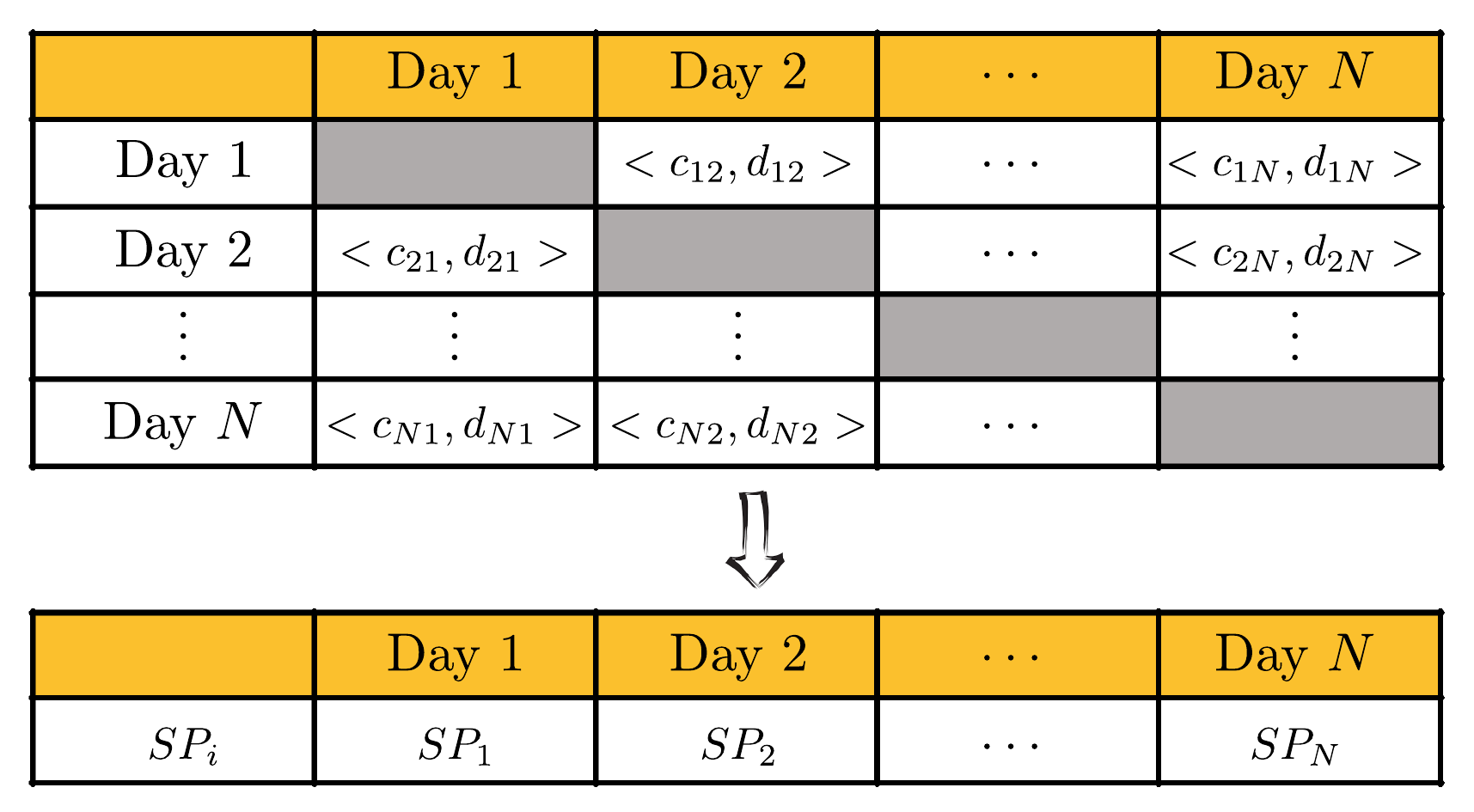}
\caption{The proposed Similarity Profile (SP)}
\label{fig_SP_matrix}
\end{figure}

Another key improvement in the proposed RM method is the capability to find the most repeated sub-patterns in the long time series instead of the approximated one. MP-based methods save the similarity value of a sub-pattern to its nearest neighbour and extract the sub-patterns with the largest similarity value as the motif, namely the most similar sub-pattern pairs in the long time series. To find the most repeated sub-patterns, however, we propose a Similarity Profile (SP), shown in Fig \ref{fig_SP_matrix}, which counts the number of similar patterns in the time series to the current sub-pattern (query pattern) as well as the average distance. Notably, a minimum of three days of data is needed when the distance is measured between pattern pairs, i.e., $N\geq 3$, since discovering what day is more repeated in two days worth of data is meaningless. 
When two sub-patterns are repeated the same number of times, we use the normalised average similarity value to choose the dominant sub-pattern. A pattern is assumed similar when its annotated DTW distance to the query pattern is smaller than a predefined threshold, $T$. The process can be presented by:
\begin{align}\label{Counter_equation}
    &c_{i,j} = 
  \begin{cases}
    1, \;\;d_{i,j}\leq T\\
    0, \;\;d_{i,j} > T
    \end{cases}
\end{align}
\noindent where $c_{i,j}$ is the similarity index, which is 1 when day $i$ and day $j$ have a small annotated DTW, $d_{i,j}$, smaller than the $T$. The threshold can be set dynamically, e.g., using the median of all annotated DTW distances in the table, or simply a fixed educated/expert guess. 
We prefer the fixed threshold approach for RM discovery in this paper to preserve the low complexity and a uniform standard for similarity measurements. 

After populating the table for all $N$ days, we can find the average annotated DTW for each day, $\hat{d_i}$, by
\begin{equation}
\label{eq:d_hat}
    \hat{d_i} = \frac{\sum_{j=1}^Nd_{i,j}}{N-1}
\end{equation}

Also, we can calculate the total similarity indices, i.e., $\hat{c_i}$, for the day $i$ as
\begin{equation}
    \label{eq:c_hat}
    \hat{c_i} = \sum_{j=1}^N c_{i,j}
\end{equation}

Finally, the SP for each day, $\text{SP}_i$, can be obtained by considering the impact of total similarity indices and average annotated DTW for the day $i$ as
\begin{equation}
    \label{eq:similarity_profile}
    \text{SP}_{i} = \hat{c_i} - \frac{\hat{d_i}}{\max(d_{i,j})}
\end{equation}
In Equation~\eqref{eq:similarity_profile}, the total similarity indices $\hat{c_i}$ are integers representing the number of similar patterns with current day $i$. The average annotated DTW distance $\hat{d_i}$ is normalised by $\max(d_{i,j})$ to be the secondary impact factor for SP, which will differentiate the days with the same similarity indices. Therefore, the proposed SP can provide both similarity indices (integral part) and average similarity (fractional part) of every daily pattern, and the largest value suggests the best motif. Algorithm~\ref{alg:SP} presents a step-by-step summary of the proposed motif discovery method.
\begin{algorithm}[!ht]
	\caption{RM discovery with SP} 
	\label{alg:SP}
	\begin{algorithmic}[1]
	    \Require {variable $pattern$ has length $n=N\cdot m$, with $N$ daily pattern $pattern_{[1:N]}$ and each daily pattern has length $m$. Pre-defined threshold $T$}
	    \State $\text{SP} \leftarrow \underbrace{[0,0...0]}_{N}$\Comment{initialise SP}
	    \State $\hat{c} \leftarrow \underbrace{[0,0...0]}_{N}$\Comment{initialise similarity indices}
	    \State $\hat{d} \leftarrow \underbrace{[0,0...0]}_{N}$\Comment{initialise average distance}
	    \State $d_{max} \leftarrow -\infty$\Comment{initialise maximum distance}
		\For {$i=1,2,3\ldots N$}
		    \For {$j=1,2,3\ldots N$}
		    \State $d \leftarrow \text{DTW}(pattern_{[i]}, pattern_{[j]})$\Comment{equations~\eqref{eq:rDTW}-\eqref{eq:mask2}}
		    \State $\hat{d}_{[i]}\leftarrow \hat{d}_{[i]}+d$\Comment{accumulate distance}
		    \If {$d>d_{max}$}\Comment{update maximum distance}
			    \State $d_{max}\leftarrow d$
			\EndIf
			\If {$d\leq T$}\Comment{check threshold}
			    \State $\hat{c}_{[i]} \leftarrow \hat{c}_{[i]} + 1$
			\EndIf
			\EndFor
			\State $\hat{d}_{[i]}\leftarrow \frac{\hat{d}_{[i]}}{N-1}$ \Comment{average distance}
		\EndFor
		\If{$d_{max}>0$}
		    \State $\text{SP} \leftarrow \hat{c} - \frac{\hat{d}}{d_{max}}$
		\Else
		    \State $\text{SP} \leftarrow \hat{c}$
		\EndIf 
		\State {$\text{RM} \leftarrow patterns_{[i]}$ \;where $\text{SP}_{[i]} == max({\text{SP}})$}
		\State \Return $\text{RM}$\Comment{RM is the Refined Motif of time series variable $pattern$}
	\end{algorithmic}
\end{algorithm}

\begin{figure}[!ht]
\centering
\begin{subfigure}{0.7\textwidth}
\includegraphics[clip, trim=0.2cm .2cm .1cm 2.2cm, width=3.6in]{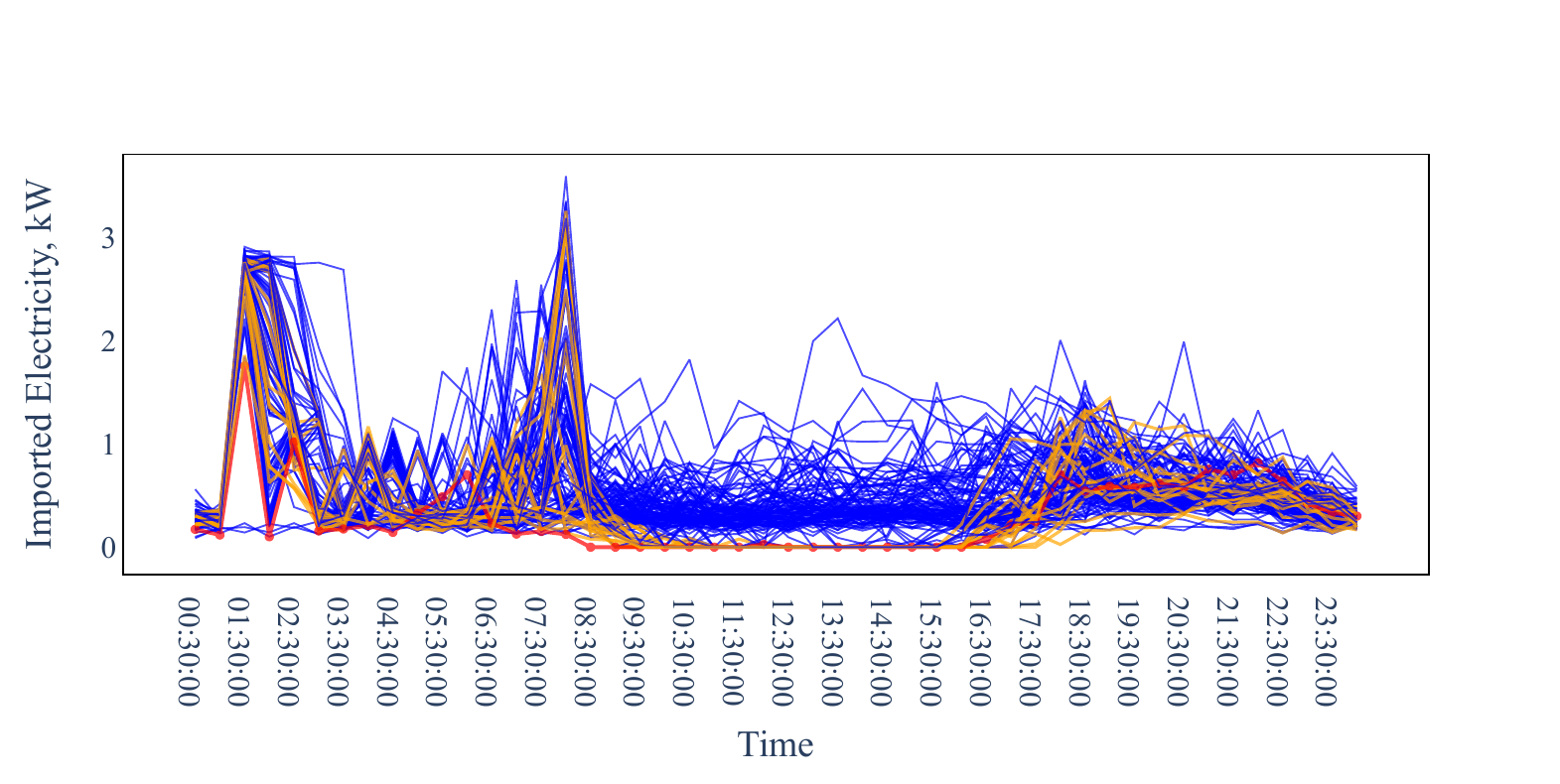}
\caption{Daily patterns with fixed threshold}
\label{fig:daily_pattern_Fixed}
\end{subfigure}
\begin{subfigure}{.7\textwidth}
\includegraphics[clip, trim=0.2cm .2cm .1cm 2.2cm, width=3.6in]{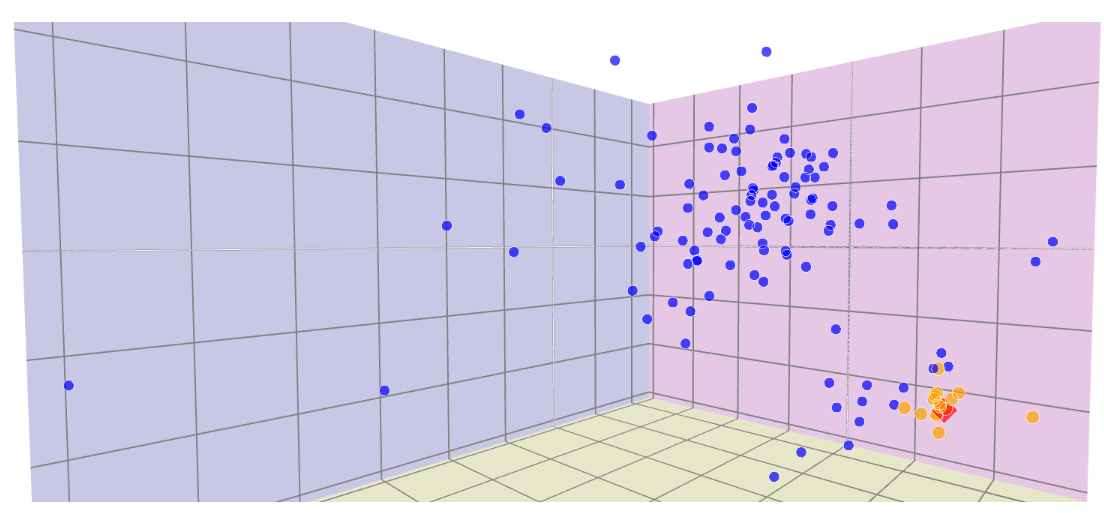}
\caption{Multi-dimensional scaling (MDS) plot with fixed threshold}
\label{fig:MDS_Fixed}
\end{subfigure}
\caption{RM (red) with the similar patterns (orange) and dissimilar patterns (blue) under \textit{fixed threshold}}
\label{RM-result}
\end{figure}

\begin{figure}[!ht]
\centering
\begin{subfigure}{0.7\textwidth}
\includegraphics[clip, trim=0.2cm .2cm .1cm 2.2cm, width=3.6in]{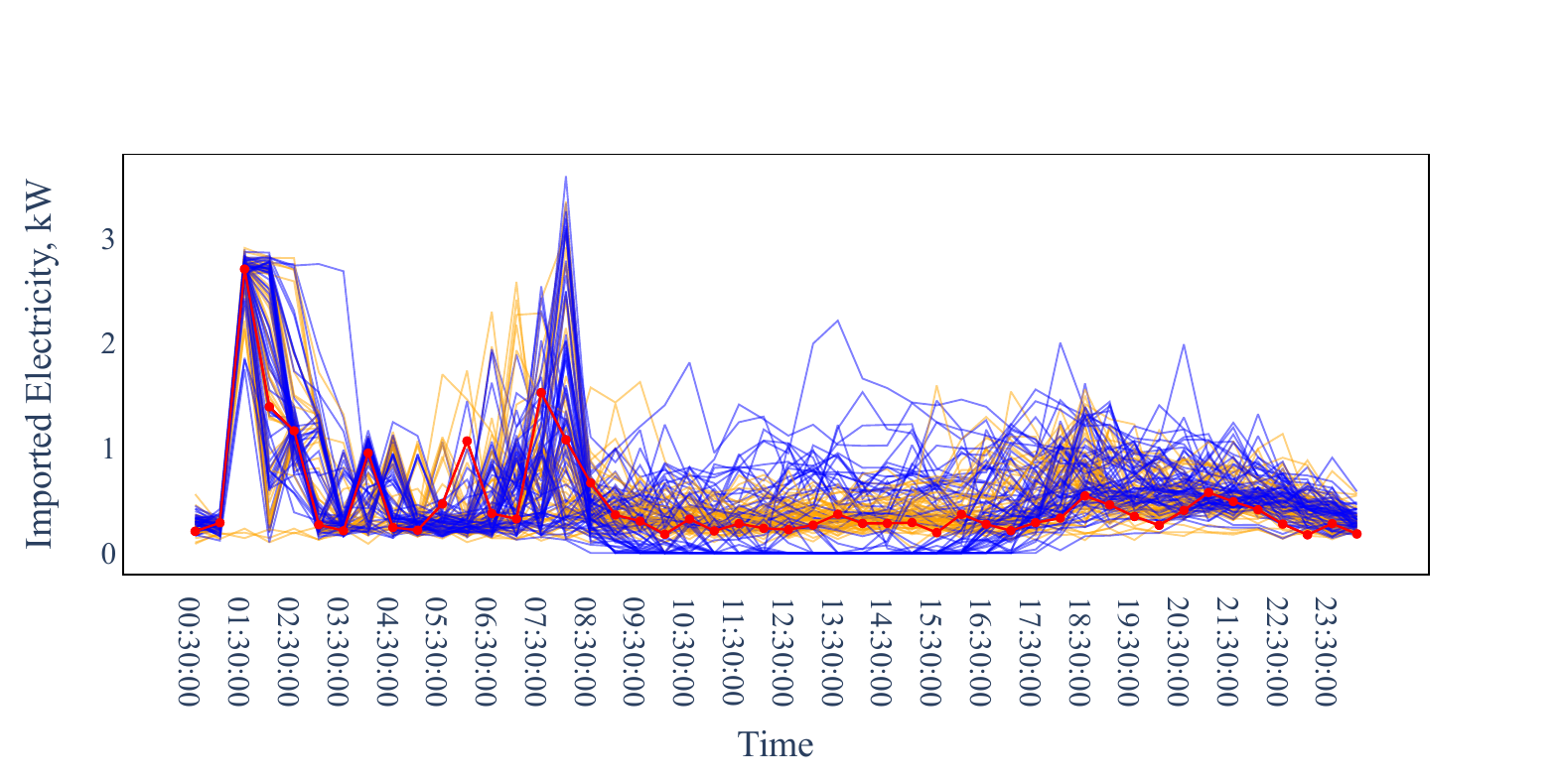}
\caption{Daily pattern with dynamic threshold}
\label{fig:daily_pattern_dyn}
\end{subfigure}
\begin{subfigure}{0.7\textwidth}
\includegraphics[clip, trim=0.2cm .2cm .1cm 2.2cm, width=3.1in]{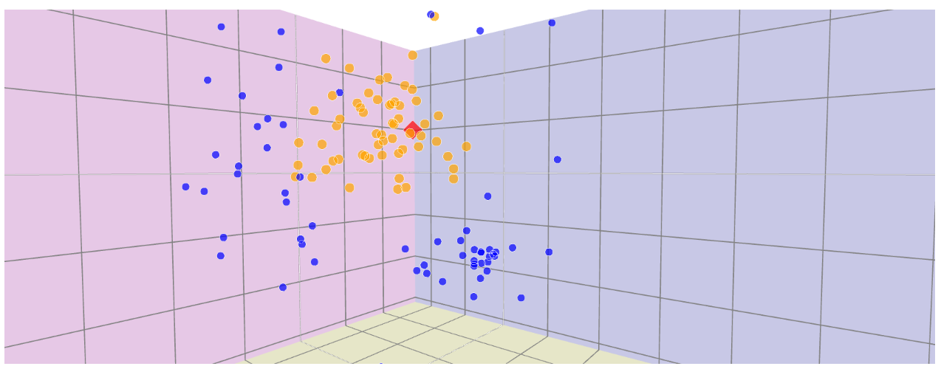}
\caption{Multi-dimensional scaling (MDS) plot with dynamic threshold}
\label{fig:MDS_dyn}
\end{subfigure}
\caption{RM (red) with the similar patterns (orange) and dissimilar patterns (blue) under \textit{dynamic threshold}}
\label{RM-result-dynamic}
\end{figure}

To show the effectiveness of the fixed threshold approach, one example is presented here for a user from our dataset, introduced in Section~\ref{experiment}. This user was a non-solar user until some time during the data collection period when the user installed a rooftop PV system. This is a challenging classification problem when the annual recorded data is used to identify the type of user. We used annotated DTW and SP, once with a fixed threshold and then with a dynamic threshold, to extract the RM for this user. The fixed threshold of 0.8 can be set with the help of domain knowledge by considering the sub-pattern length and scale. The dynamic threshold is the median of annotated DTW values. In Figures~\ref{fig:daily_pattern_Fixed} and \ref{fig:daily_pattern_dyn}, we show the daily patterns of this user using fixed and dynamic thresholds, respectively, where the RM and similar patterns are coloured in red and orange, respectively. Also, the annotated DTW distance matrix of each daily profile is mapped into a 3-D space through Multi-dimensional scaling (MDS) in Figure~\ref{fig:MDS_Fixed} and \ref{fig:MDS_dyn} for the fixed and dynamic thresholds, respectively. It can be seen from Figure~\ref{fig:daily_pattern_Fixed} that the RM and similar patterns obtained by the fixed threshold match our understanding of the PV operation by following the well-known duck curve and have small distances to the RM in 3D space in Figure~\ref{fig:MDS_Fixed}. On the other hand, in Figure~\ref{RM-result-dynamic}, the RM identified by the dynamic threshold is heavily influenced by the patterns from before the rooftop solar installation. The similar patterns obtained by dynamic threshold do not share the similar shape in Figure~\ref{fig:daily_pattern_dyn} and have large distances to the RM in Figure~\ref{fig:MDS_dyn}. As a result, we will mislabel this user as a non-solar user when using the dynamic threshold. 

\subsection{Motif-based Linear Classification Method}\label{ML}
To use the extracted RMs as classifiers, a simple Neural Network (NN) model is used to classify the users while preserving the interpretability of the Interpretable Refined Motifs in Binary Classification (IRMAC) solution. The proposed classifier contains a single neuron with a linear function in the hidden layer as in Equation~\eqref{eq:linear}, cascaded with a sigmoid function in the output layer in Equation~\eqref{eq:sigmoid}, where $\mathbf{I}$ is the input data with $k$ users' motifs of length $m$, and $\vec{a_1}$, $b_1$, $a_2$, $b_2$ are the weights and biases of the linear function and the sigmoid function, $\vec{U}_{sn}$ is the output vector from the linear function with length $k$, and vector $\vec{U}_{sig}$ with length $k$ is the sigmoid classification outputs for $k$ users: 
\begin{align}
    \vec{U}_{sn} &= \mathbf{I}\cdot\vec{a}_1+b_1 \label{eq:linear}\\ 
    \vec{U}_{sig} &= \frac{1}{1+e^{-(\vec{U}_{sn}\cdot {a_2}+b_2)}} \label{eq:sigmoid}
\end{align}

Hence, by Equations~\eqref{eq:linear} and \eqref{eq:sigmoid}, the classification problem can then be represented as
\begin{align}
\label{eq:classification_final}
    \vec{U}_{sig} = \frac{1}{1+e^{-(\mathbf{I}\cdot \vec{a}_1 \times a_2 + a_2\times b_1 + b_2)}}
\end{align}

Since the sigmoid function is an increasing function, and $a_2$, $b_1$, and $b_2$ are constant scalars and biases that do not affect the comparison relationships among users, Equation~\eqref{eq:classification_final} can be further simplified to $\mathbf{I}\cdot\vec{a}_{1}$, which has linear complexity, $O(k)$, where $k$ is the number of users. In addition, this weight vector has values that align with our domain knowledge (hence interpretability), which will be further discussed in Section \ref{experiment}.

\subsection{Further Advantages of the Proposed Method}
\label{sec: systematic advantages}
\begin{figure}[!ht]
\centering
\includegraphics[clip, trim=2cm 1cm 2cm 1cm, width=3in]{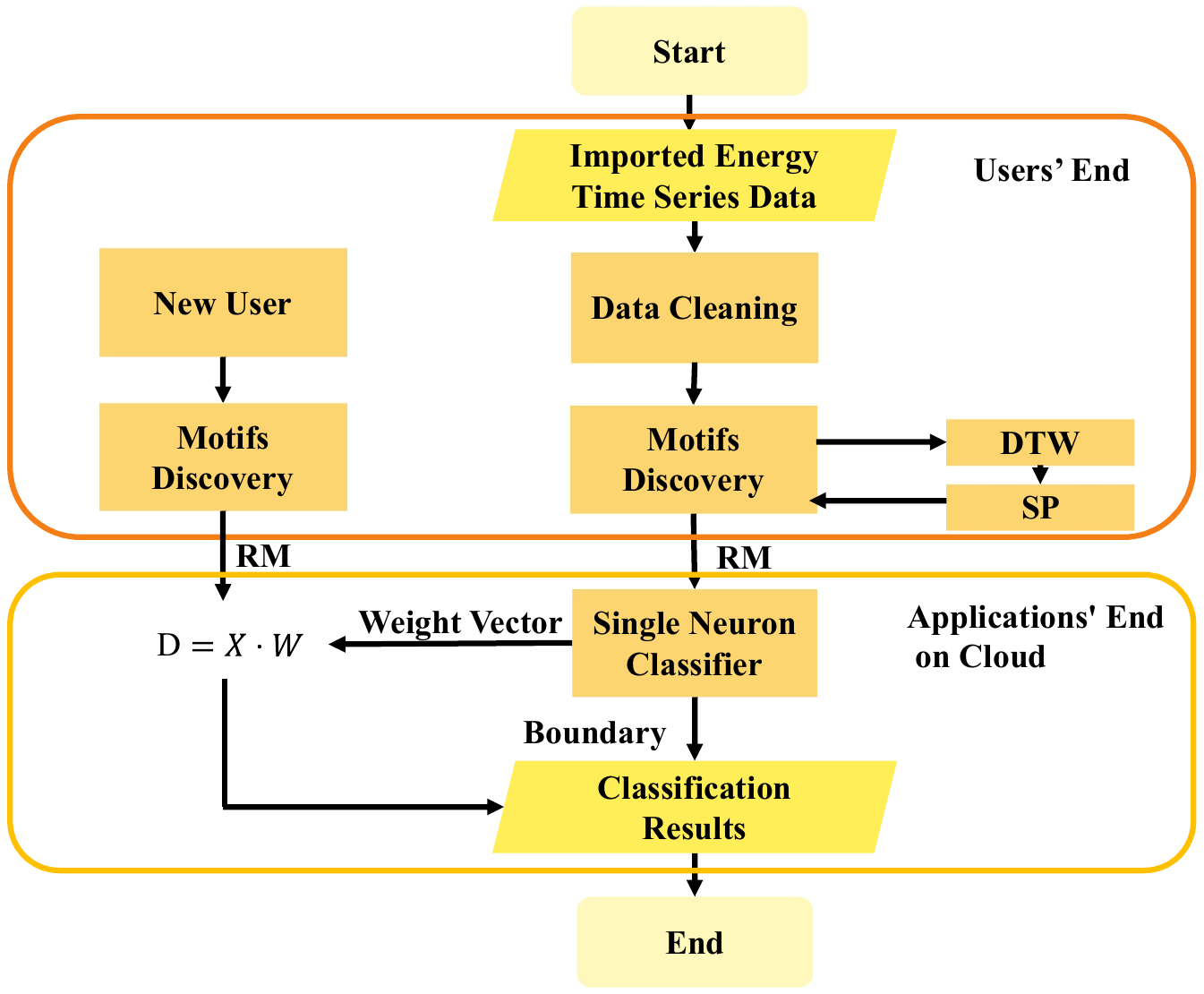}
\caption{Block diagram for the proposed IRMAC method}
\label{fig_sim}
\end{figure}

The proposed classification method can be implemented as in Figure \ref{fig_sim}. 
The imported energy time series will be cleansed in the pre-processing stage before applying annotated DTW to obtain the SP. The extracted motif of each user is the input to the single neuron classifier. The trained model is a linear function used for new user classification.

Furthermore, the proposed method can be implemented hierarchically where the end-users at the edge of the grid can extract motifs at their end. As a result, seven key advantages can be expected using the proposed method, as follows.

\begin{enumerate}
    \item It is memory efficient since only the extracted motif needs to be stored instead of the long time series of consumption data for each user.
    \item It needs low communication bandwidth as the end users communicate the motif to the cloud system instead of the whole time series.
    \item The fast processing speed can be achieved since motif discovery is conducted at the users' ends in parallel, whereas the classification method is a linear function taking only motifs as inputs.
    \item The cost for the central entity, e.g., aggregator, can significantly decrease because of the low storage and computation capacity requirement.
    \item Moderate abnormal patterns of end users cannot influence the performance of the proposed method since only motifs can be extracted and used for classification purposes.
    \item The proposed method lowers privacy concerns regarding data leakage and is robust to false or missing data.
    \item The proposed RM method is easily maintainable.
\end{enumerate}

As a further explanation of the sixth advantage, please note that the high-resolution consumption data is considered sensitive as one can easily identify the household occupancy, lifestyle, and the usage of different appliances from that \cite{Briggs2020}. With the proposed method, however, only the motif, i.e. one day's partial data, for each user is communicated to the aggregator and recorded in their databases. In case of data leakage, only one day's worth of RM data, or less than one day of data when only certain periods of the day is of interest, is exposed. This is not a significant privacy threat because the extracted RMs do not contain the full information about the end-user's lifestyle and occupancy, and the users' data identification model is not at the user's end. Also, the proposed method is more robust to false or missing data at the users' ends than ML-based methods since those abnormal behaviours occur irregularly and will not affect the most repeated patterns, i.e. RMs. Additionally, the false data of one user cannot affect other users because RM discovery for each end user is highly independent, which is not the case in some ML-based methods, e.g., federated learning \cite{Wu2022}.

To better clarify what we mean by `maintainability' in the last item above, consider a case where new users' data must be included. In this case, updating the model to include new users' data only requires the motif from the new user rather than recomputing previous motifs for all consumers and refining the models in federated-based learning models \cite{Wus2021}. Therefore, we do not need to repeat the entire motif discovery process for a user to update its motif. In summary, the proposed method is fast, computationally efficient, secure from the end-users perspective, scalable and interpretable.

\section{Simulation Results}
\label{experiment}
To show the applicability of the proposed IRMAC method, the experiments are conducted on two benchmark datasets to solve two different binary classification problems: 1) finding rooftop solar PV among residential consumers, and 2) finding residential consumers with electrical heating systems. In the first dataset, we have half-hourly PV generation and load demand data of 300 residential consumers from the East Coast of Australia \cite{datasourcewebsite}. The database contains one year's worth of gross solar PV generation, general electricity consumption, and electricity consumption profile of a controlled appliance (likely a hot water system), which leads to an electricity usage surge around midnight. We also know the capacity of the PV panels in each household. 
We have done a quality check on the dataset and divided it into imported energy from the grid for 300 solar and 300 non-solar users. The simulations were run on a Windows machine with an AMD\textsuperscript{\textregistered} Ryzen 5 6-core processor and RTX 2060 using \texttt{Python} 3.8.

Secondly, the imported energy data used in the electric heating system identification problem is obtained from our industry partner \cite{watts}. The dataset contains 10,000 residential consumers from across Denmark. After data pre-processing, we picked 1052 consumers in the same region. The sample dataset had 893 users with no electric heating system and 159 users with electric heating systems. It is an unbalanced dataset for modelling purposes. To analyse this unbalanced data, the F-score is used as a performance measure, a widely accepted measure of balance between the precision and the recall in unbalanced datasets for binary classification \cite{Goos2019}. We use the flat F-score instead of $F_\beta$-score in this paper since only a classification task is discussed. 

We applied out-of-sample validation on the two datasets introduced above. For a fair comparison, all methods are built and tested based on the same training \& testing sets. The specific train \& test set arrangement is discussed in Section \ref{sim:PV} and \ref{sim:EH}, respectively. 

Another performance measured in the simulation results is interpretability. With many papers assessing interpretability differently, this paper tries to use the most applicable discourse, that is 
\begin{quote}
    A classification method is ``interpretable" if the algorithms are transparent and fully understandable to the people who employ it \cite{Zhang2021, Kamath2021, Guidotti2018}.
\end{quote}

\subsection{Solar PV Users Identification}\label{sim:PV}
\begin{figure}[!ht]
\centering
\includegraphics[clip, trim=0cm 0cm 0cm 0cm, width=5in]{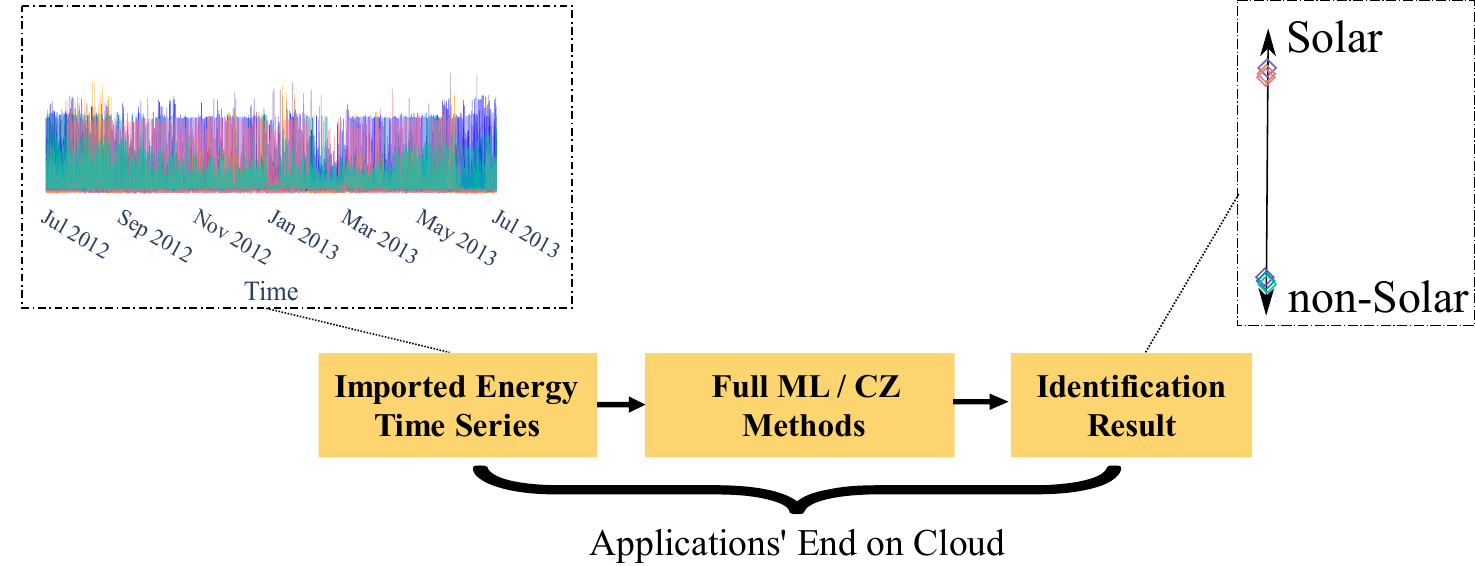}
\caption{A schematic showing the processing steps of the Full ML and CZ methods with 6 sample users}
\label{fig_block_overall_ML}
\end{figure}

\begin{figure}[!ht]
\centering
\includegraphics[clip, trim=0.1cm 0.1cm 1.8cm 1.56cm,width=0.49\textwidth]{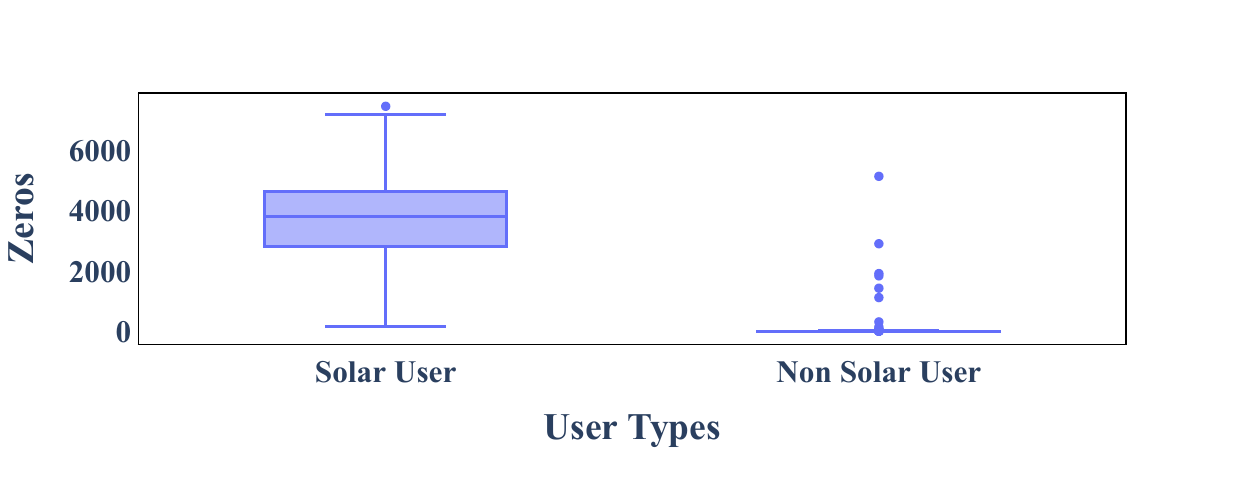}
\caption{Box plot for the ``Counting Zeros (CZ)'' method}
\label{box-plot}
\end{figure}

Based on Occam's razor theory, sophisticated methods make sense only when a problem cannot be solved with a direct or intuitive solution. As a result, we compare the proposed RM method with several intuitive and complex methods as follows:

\begin{enumerate}[leftmargin=*]
    \item Full ML:
    \begin{enumerate}
        \item Two Classic Conventional ML Classification Methods:
    \begin{itemize}
        \item Support vector machines (SVM): We use the C-Support Vector Classification. It is implemented with \texttt{sklearn}, where RBF (Gaussian) kernel is used following \cite{chang2011libsvm}.
        \item K-Nearest Neighbors algorithm (k-NN): We set the number of neighbours to 2, which is data dependent and chosen based on trial and error \cite{Zhang2017}.
    \end{itemize}
    \item Two DNN Methods:
    \begin{itemize}
        \item Multi-layer perceptron (MLP): A DNN with $5$ layers sequential model containing ReLUs and dropout.
        \item Time Series Convolutional Neural Network (1D-CNN): A state-of-the-art DNN specifically designed for time series classification, which outperforms other ML solutions \cite{brunel2019cnn}. The employed model contains two 1D Convolutional layers, dropout, 1D Max Pooling, one dense layer and a sigmoid output layer.  
    \end{itemize}
    \end{enumerate}
    \item Intuitive: 
    \begin{enumerate}[leftmargin=*]
        \item Counting Zeros (CZ): The general understanding is that a consumer with rooftop PV does not import energy in the middle of the day most of the time; hence resulting in numerous zero entries in the imported electricity time series. In this method, we examine this intuitive idea to identify solar PV owners. From the box plot in Figure \ref{box-plot}, we can see that most of the non-solar users in the training set have less than 2,000 zero entries, while most solar users have more than 2,000 zeros. Therefore, we can use this value to classify consumers in this approach.
        \item Average Daily Profile: To challenge the RM method, we replaced extracted motifs with the average daily profiles of each user. Similar to IRMAC, we used a single neuron classifier for the classification problem in this case.
    \end{enumerate}
    \item Motif-based methods:
    \begin{enumerate}[leftmargin=*]
        \item STOMP Motifs: It is the most dominant motif discovery method in the literature, which uses one interval step size pairwise ED with MP.
        \item DTW Motifs: Instead of ED, we employ the DTW to compute the nearest neighbour pairs. Although some research studies increased motif discovery speed by omitting some computations, we use the brute force approach for the most accurate result, as the offline motif discovery process is not a major concern when accuracy is the major concern \cite{Alaee2020,Lagun2014}.
        \item IRMAC: This is our proposed method.
    \end{enumerate}
\end{enumerate}
Both Full ML and CZ methods require the whole year of data of every consumer on the cloud as shown in the Figure \ref{fig_block_overall_ML}. On the other hand, Motif-based methods and Average Daily Profile extract one day worth of data at users' end. The data is than processed by the classifier introduced in Section \ref{ML}. 

\begin{figure}[!ht]
\centering
\includegraphics[clip, trim=0.1cm 0.1cm 1cm 2.4cm, width=3.5in]{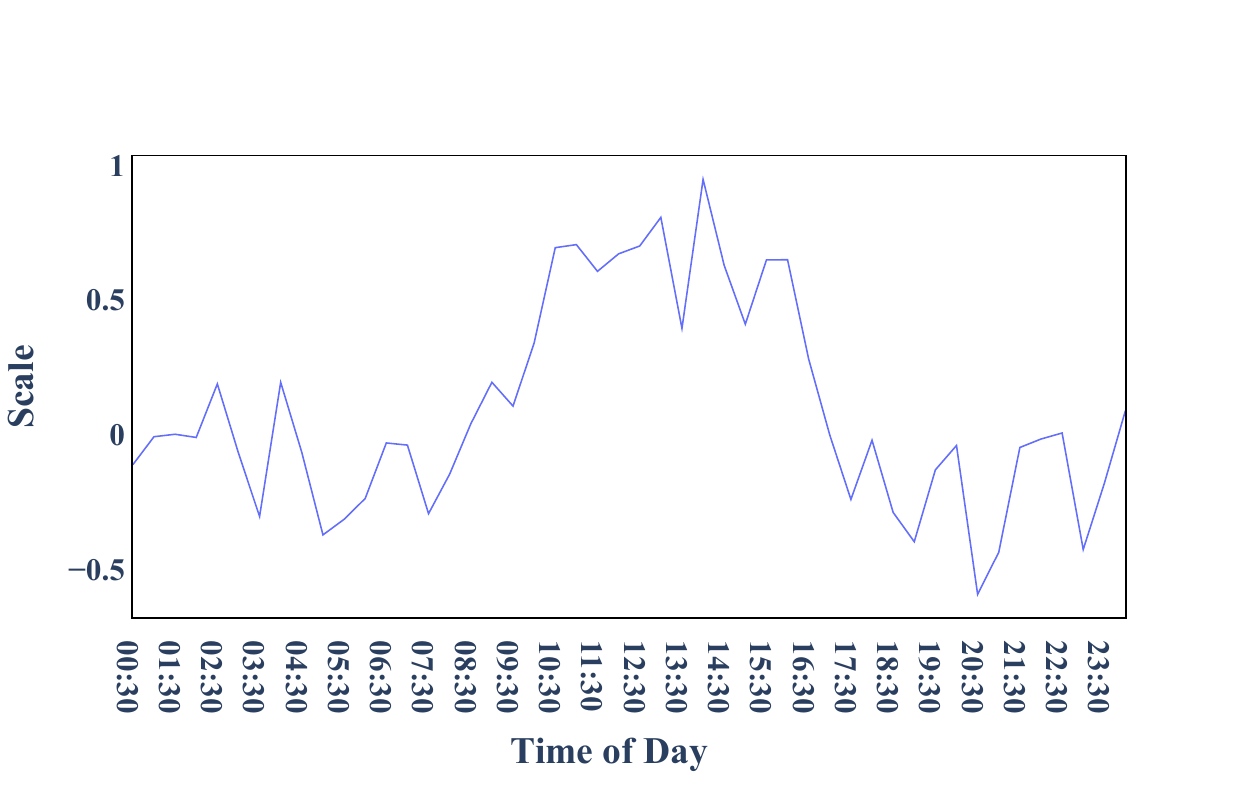}
\caption{Weight vector obtained during single neuron classifier training}
\label{weight}
\end{figure}

For the dataset from the East Coast of Australia, only $91$ days of summertime imported electricity data is used to have the strongest solar generation pattern in the time series, reducing the motif discovery time in the motifs-based methods. 
With extracted motifs from SP as inputs, we devote 448 motifs to the training set and the remaining 150 motifs to the testing set. 
The calculated weight vector from the single neuron classifier is shown in Figure \ref{weight}, which is obtained from the initial value of the weight mask set based on the sunset and sunrise times, with a $90\%$ accuracy for the testing set. The overall trend of the weight profile matches our domain knowledge of solar PV generation characteristics. This weight profile can guide us in calibrating the annotated DTW for SP by limiting the window to $10$ am to $4$ pm and using the same mask on the weight vector for classification. These calibrations improved the IRMAC's accuracy to $96\%$ on the testing dataset. Two solar users have been detected as non-solar users, i.e. False Negative (FN), and four non-solar users have been detected as solar users, i.e. False Positive (FP). 

The different classification methods are compared in Table~\ref{Comparision table:PV}. In general, the Motif-based and full ML techniques outperform the intuitive methods. The better performance of IRMAC over the Average method indicates that the extracted motifs contain significant features of the PV generation compared to the annual average values. Within the motif-based methods, STOMP and DTW-based motifs are incapable of finding PV features for three reasons. First, early morning and evening load patterns are dominant in a daily profile for most consumers; thus, the most significant features are not from solar generation. Second, the classic STOMP motif discovery uses a single-step sliding window to include all possible sub-patterns with overlaps, which mixes the nighttime patterns with the distance counting. Therefore, the extracted motifs failed to be classified by a linear function. Finally, both the STOMP and the DTW-based motifs look for the pair of sub-patterns with the highest similarity, which tends to be affected by special situations that scarcely happen, e.g., users go out for vacations or users have guests for two days. 

By comparing the accuracy and F-score in Table~\ref{Comparision table:PV}, the 1D-CNN outperforms other methods. However, it is an opaque method and hence not interpretable. Furthermore, it requires 335s during training to get the reported accuracy, compared to 125s for MLP and 35s for IRMAC. Also, 1D-CNN has a higher classification time, as in Table~\ref{Comparision table:PV}. Note that the offline training time and motif discovery computation time are not reported in the table. This is because the models are not required to be retrained regularly, and motif discovery is processed in parallel for each user at their end; thus, the actual classification time in the IRMAC method is only the time needed for the linear classification function. The motif discovery will be carried out once for each user, and it takes about 20s from summertime data using the regular compiler. After that, we need only to update the motif by incoming daily data, which requires 300ms for each user on average. Therefore, the motif discovery time should be separated from the actual classification problem, which takes 1.03 ms to solve. To fairly compare the motif discovery time of IRMAC, DTW-based motif and STOMP, we used the \texttt{Numba} compiler for the three methods. It took IRMAC 130ms for a single user's motif discovery, compared to 40ms in STOMP and 180s in DTW-based motif, which shows the complexity of the proposed RM discovery and STOMP matches the theoretical values given in Section \ref{method}. As the Average and STOMP methods used the same linear classification method proposed as in IRMAC, the three are reported with the same speed. 

In conclusion, the proposed IRMAC method outperforms the other techniques considering the computational time, accuracy, and transparency. The advantage of computational speed will become increasingly important in real-world applications where many users' data need to be processed centrally. In addition, when considering iterative data exchange through communication links in some ML-based approaches (like federated learning), the advantages of our proposed method will be better shown.
\begin{table}[htb]
    \centering
    \caption{Performance of different methods in PV owner classification problem}
    \resizebox{1\textwidth}{!}{
    \begin{tabular}{*{10}{c}}
        \toprule
        \rowcolor{TBLHeader} & \multicolumn{4}{c}{Full ML} & \multicolumn{2}{c}{Intuitive} & \multicolumn{3}{c}{Motif-based} \\\cmidrule(l){2-5}\cmidrule(l){6-7}\cmidrule(l){8-10}
        \rowcolor{TBLHeader} & SVM & KNN & 1D-CNN & MLP & Average & CZ & STOMP & DTW motif & \textbf{IRMAC} \\\midrule
         Accuracy & 89.4\% & 72\% & \textbf{97.3\%} & 94.6\% & 87\% & 94.6\% & 56\% & 61\% & 96\% \\ 
         \rowcolor{TBLRow}F-score & 0.897 & 0.71 & \textbf{0.973} & 0.945 & 0.86 & 0.944 & 0.56 & 0.64 & 0.96\\
         Time, s & 1.93 & 0.5 & 1.2 & 1.1 & \textbf{0.001} & 0.8 & \textbf{0.001} & \textbf{0.001} & \textbf{0.001} \\
         \rowcolor{TBLRow}Algorithmic transparency & \xmark & \cmark & \xmark & \xmark & \cmark & \cmark & \cmark & \cmark & \cmark \\
         \bottomrule
    \end{tabular}
    }
    \label{Comparision table:PV}
\end{table}

\begin{figure}[!ht]
\centering
\includegraphics[clip, trim=2cm 0.1cm 2cm 2cm, width=3.7in]{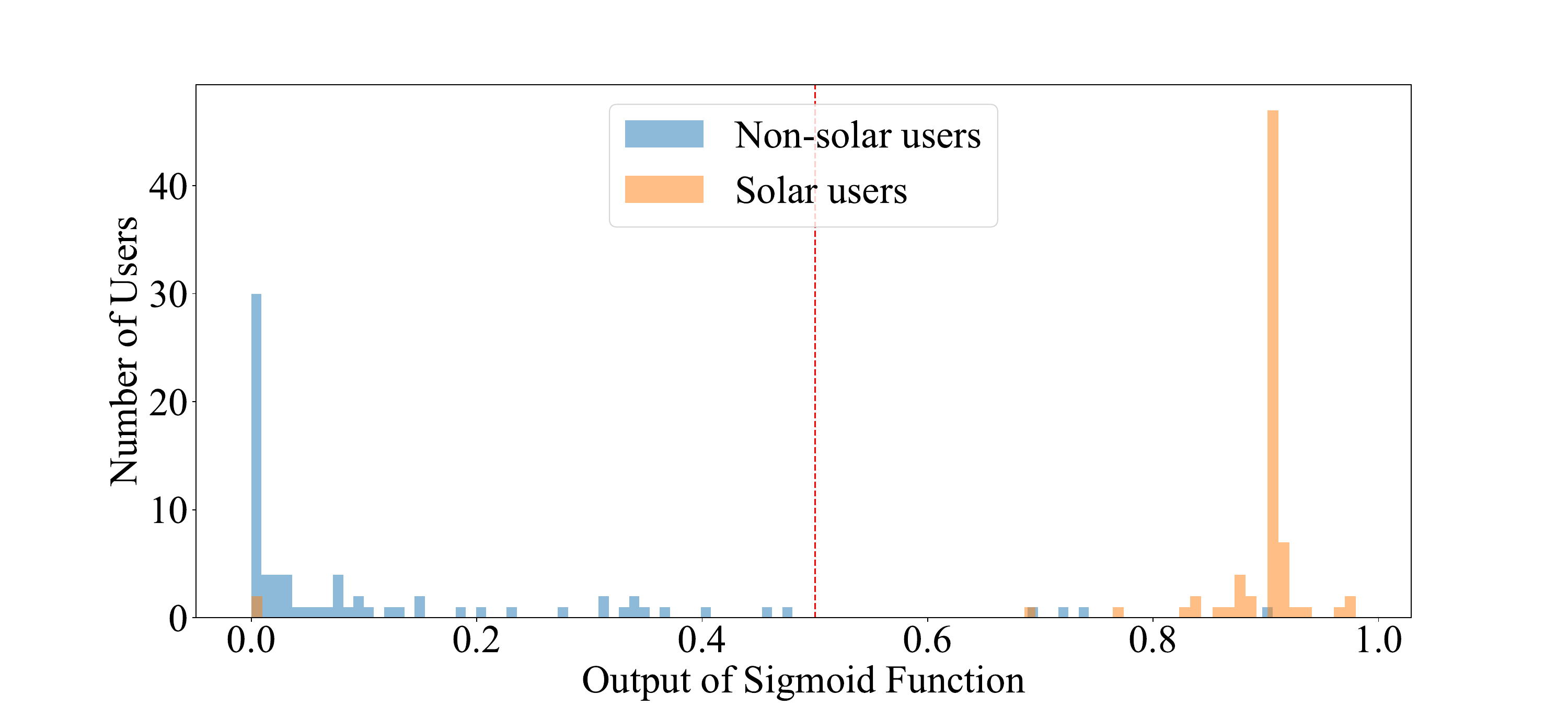}
\caption{A histogram of the IRMAC method from the testing dataset}
\label{KDE}
\end{figure}

To further study the proposed method, Figure \ref{KDE} shows the histogram of the results from the test dataset, where all results from the sigmoid function are coloured by their real user types. While the IRMAC model classified the users with a value below 0.5 as non-solar users, we can observe two FN on the far left and four FP on the right. Most of the results are located at the two ends, showing IRMAC model is likely to give a confident result. However, among the falsely detected users, two FNs and one FP are located at the far ends, which means the proposed method has high confidence in classifying them correctly. It requires further analysis as follows.

\begin{figure}[!ht]
\centering
\includegraphics[clip, trim=0cm 0.3cm 2cm 2.2cm, width=0.8\textwidth]{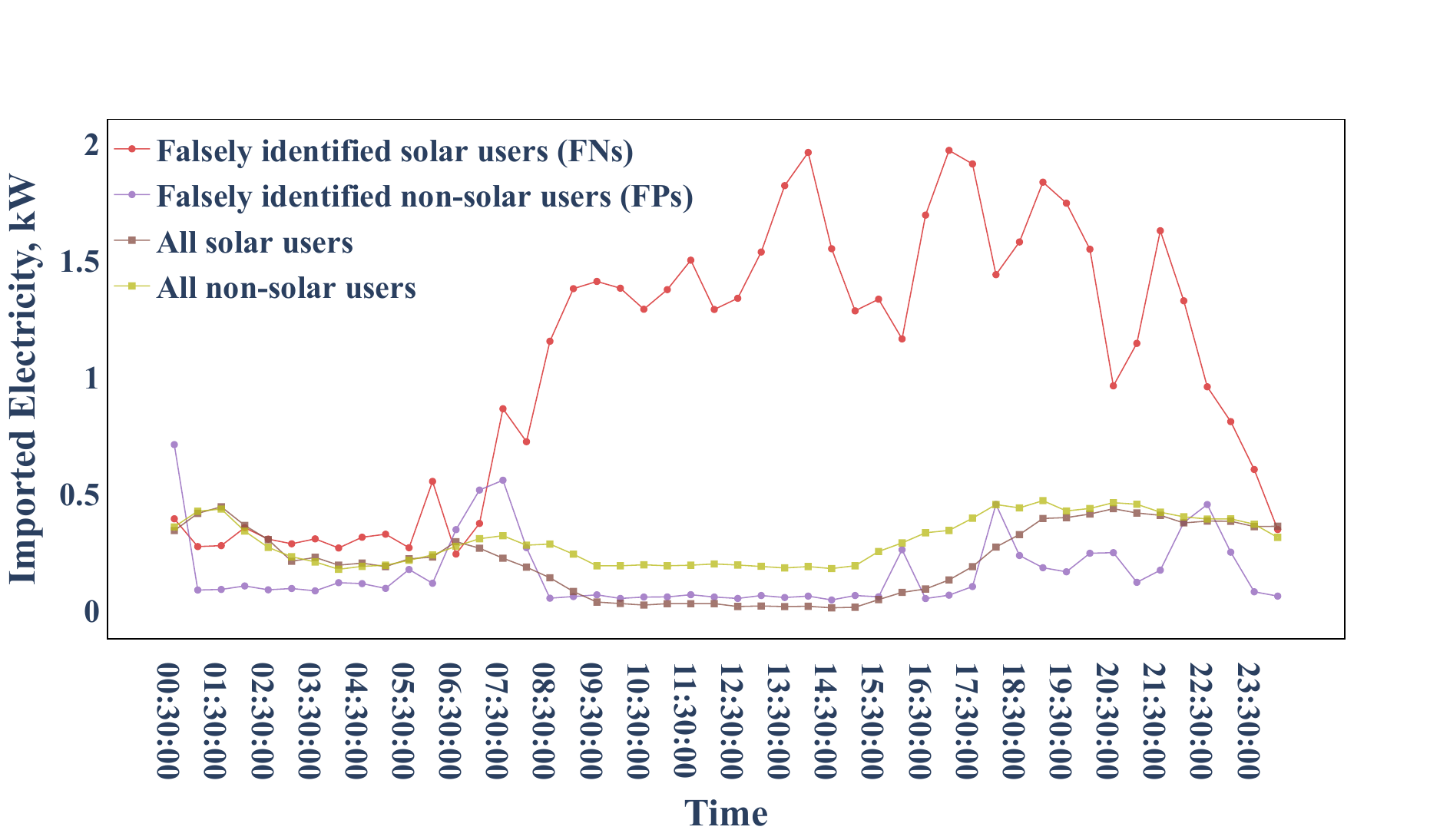}
\caption{Average motifs of different users' groups}
\label{Motifs}
\end{figure}

\begin{figure}[!ht]
\centering
\includegraphics[clip, trim=2cm 0.1cm 2cm 2cm, width=3.7in]{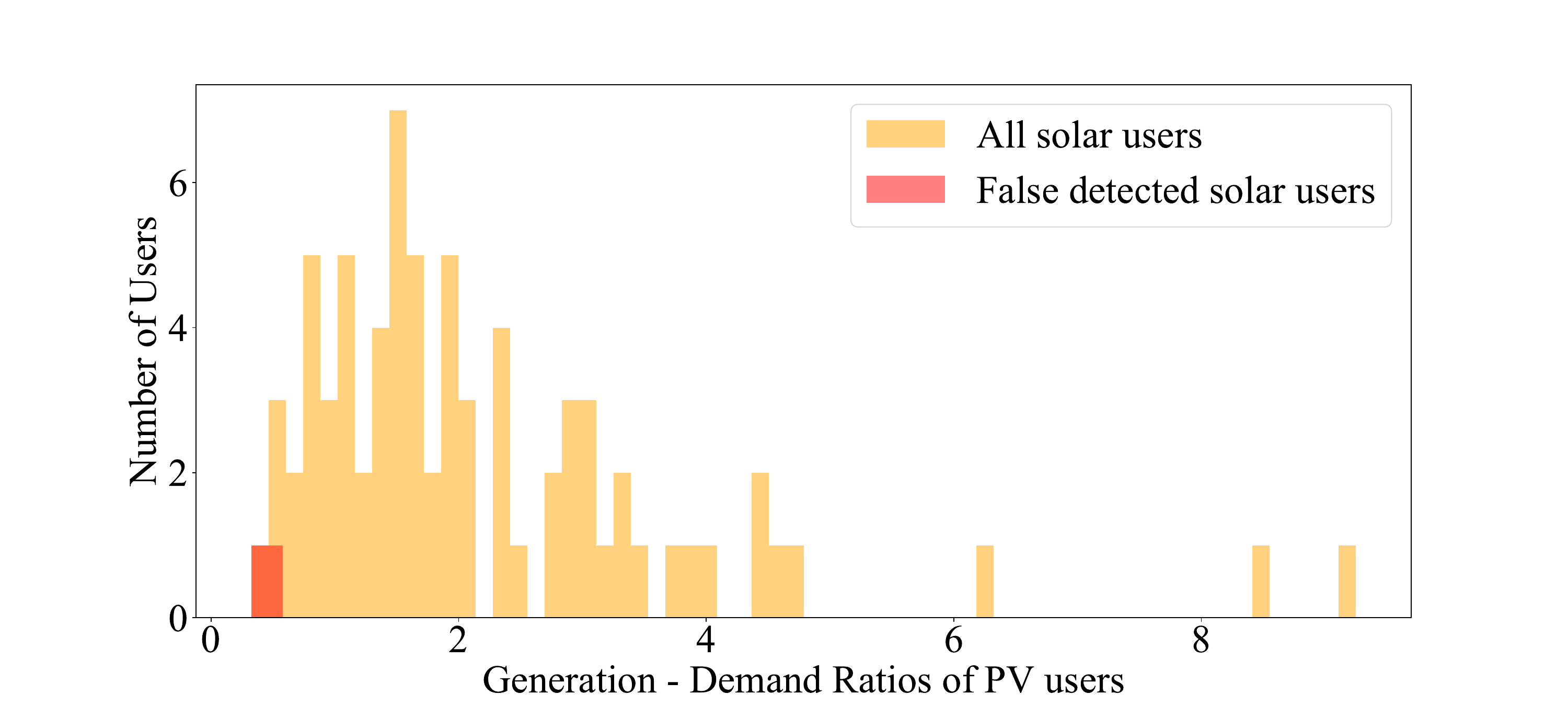}
\caption{The ratio of total PV production to the total daytime consumption for solar users}
\label{PV-Hest}
\end{figure}

First, we compare the FN and FP average motifs with the average motifs of solar and non-solar users in Figure \ref{Motifs}. We can see that the falsely detected solar users have irregular profiles compared to their respective groups. For example, while solar users have a relatively flat and close to zero motif shape during the daytime, the FN cases have significantly higher values during the same time. It means that these two FN cases have either a tiny PV system or exceptionally high demand during the daytime. This is investigated further by looking at the gross solar energy generation, demand, and solar system capacities of the two FN users. First, we noticed that the two FN users have $1.0$ kW PV systems. 
Furthermore, from the generation-consumption ratio plot in Figure \ref{PV-Hest}, we observe that the two FN users have the lowest daytime generation-demand ratio. This indicates that our method cannot identify the solar users with high electricity demand and low solar production.

To check on the performance of the proposed identification method, we separately tested the IRMAC on wintertime data when the solar generation was comparatively low. The simulation study showed an accuracy of $95.3\%$ for winter data, which is almost the same as the accuracy of $96\%$ in the summertime. Among the misidentified users, there were four solar users with no solar generation during the whole month of June (winter month in the Southern hemisphere), although they had a normal solar generation in July and August. Hence they are solar users for the whole wintertime but are non-solar users if we only use the data from June. 
This simulation study proves that the proposed solution performs accurately in different seasons. Furthermore, it shows the necessity of dynamically identifying solar users based on their imported data due to faulty solar systems and changes in consumers' behaviour over time.

\subsection{Electrical Heating System Identification}\label{sim:EH}
\begin{figure*}[!ht]
    \centering
    \subfloat[Electric heating system users in winter \label{LD:wintere}]{\includegraphics[clip, trim=0.05cm 0.6cm 1.8cm 2.6cm,width=0.49\textwidth]{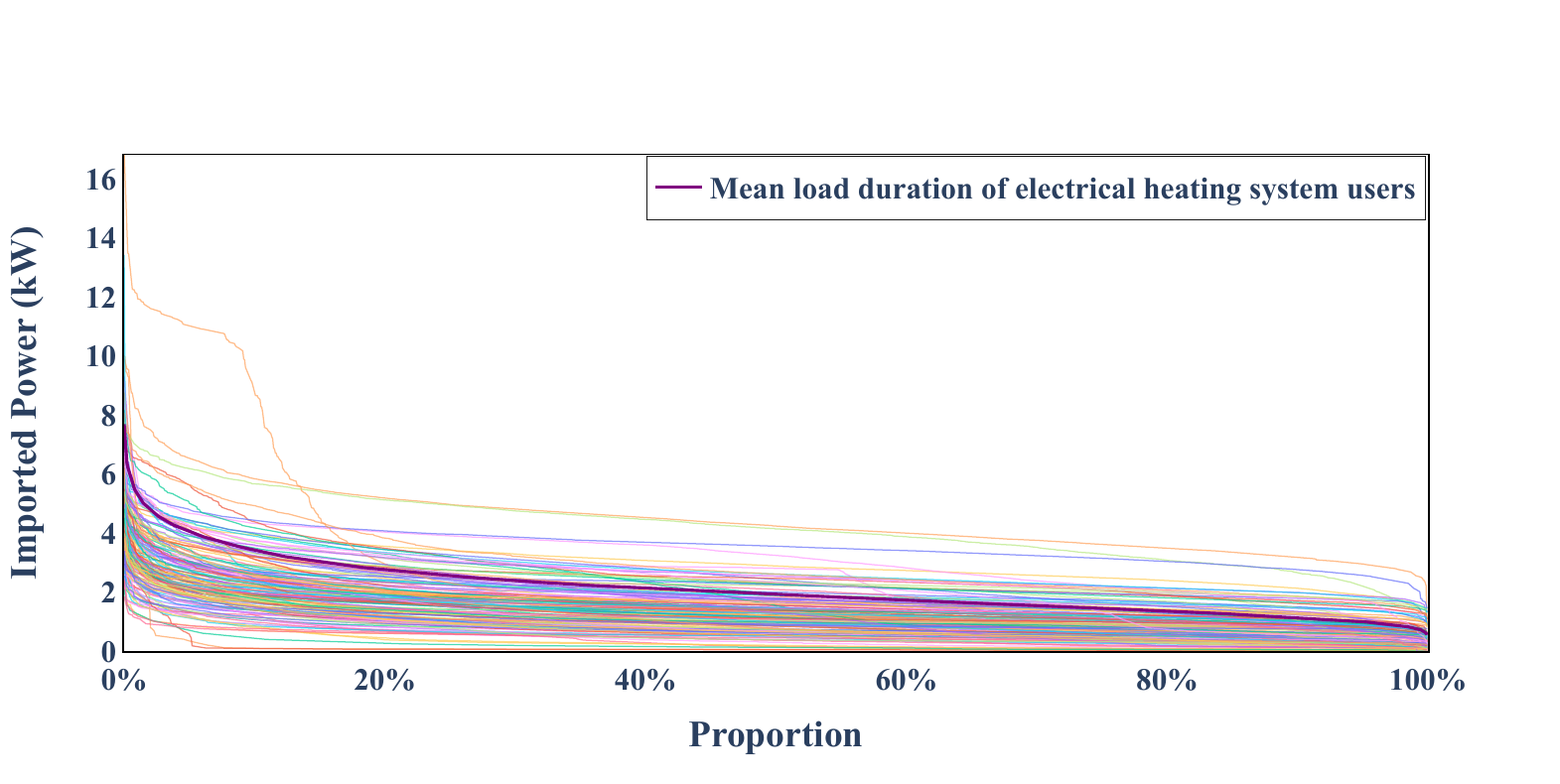}}
    \subfloat[Electric heating system users in summer\label{LD:summere}]{\includegraphics[clip,trim=0.05cm 0.6cm 1.8cm 2.6cm,width=0.49\textwidth]{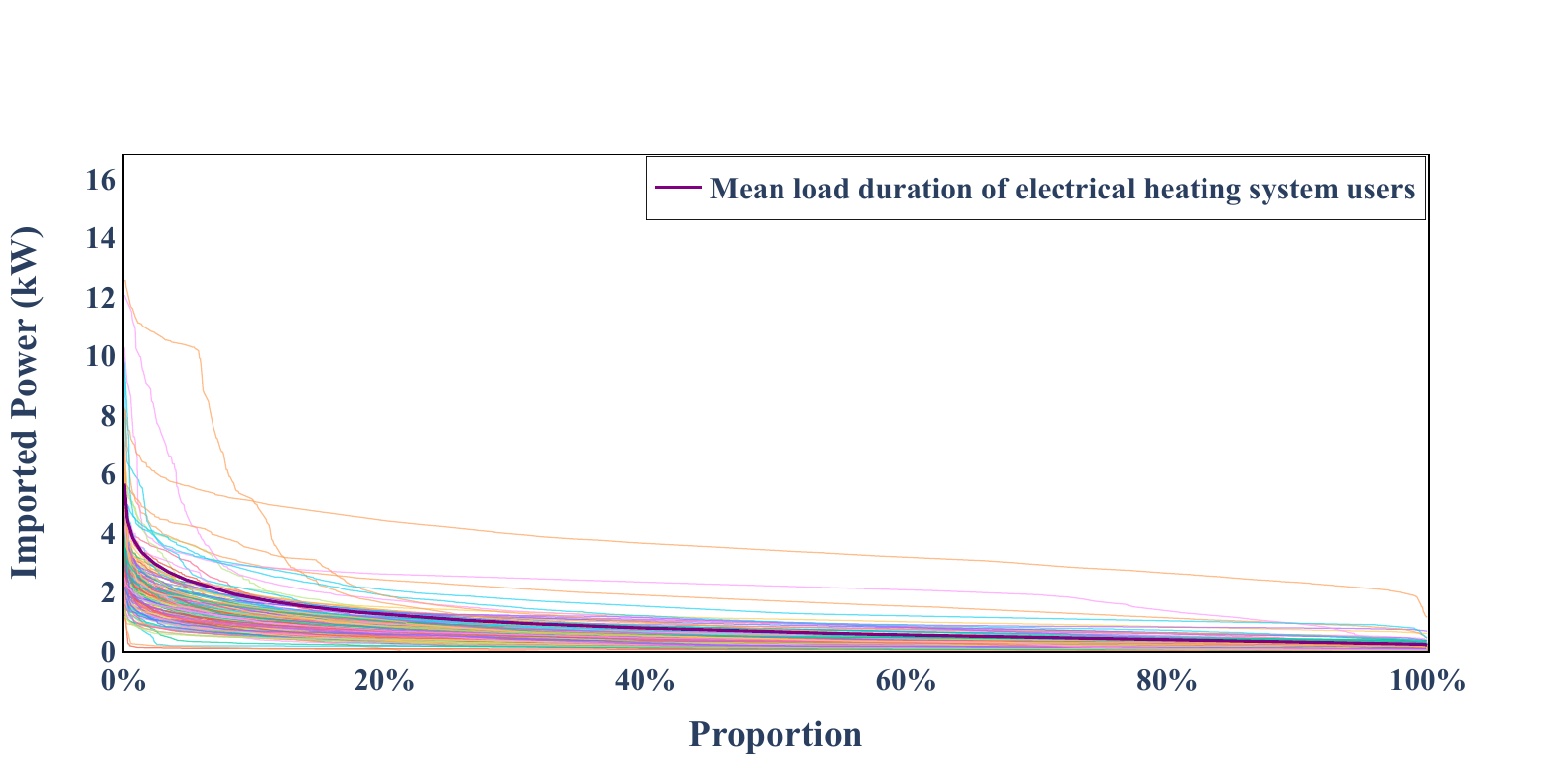}}\vskip\baselineskip
    \subfloat[Non-electric heating system users in winter\label{LD:winterne}]{\includegraphics[clip, trim=0.05cm 0.6cm 1.8cm 2.55cm,width=0.49\textwidth]{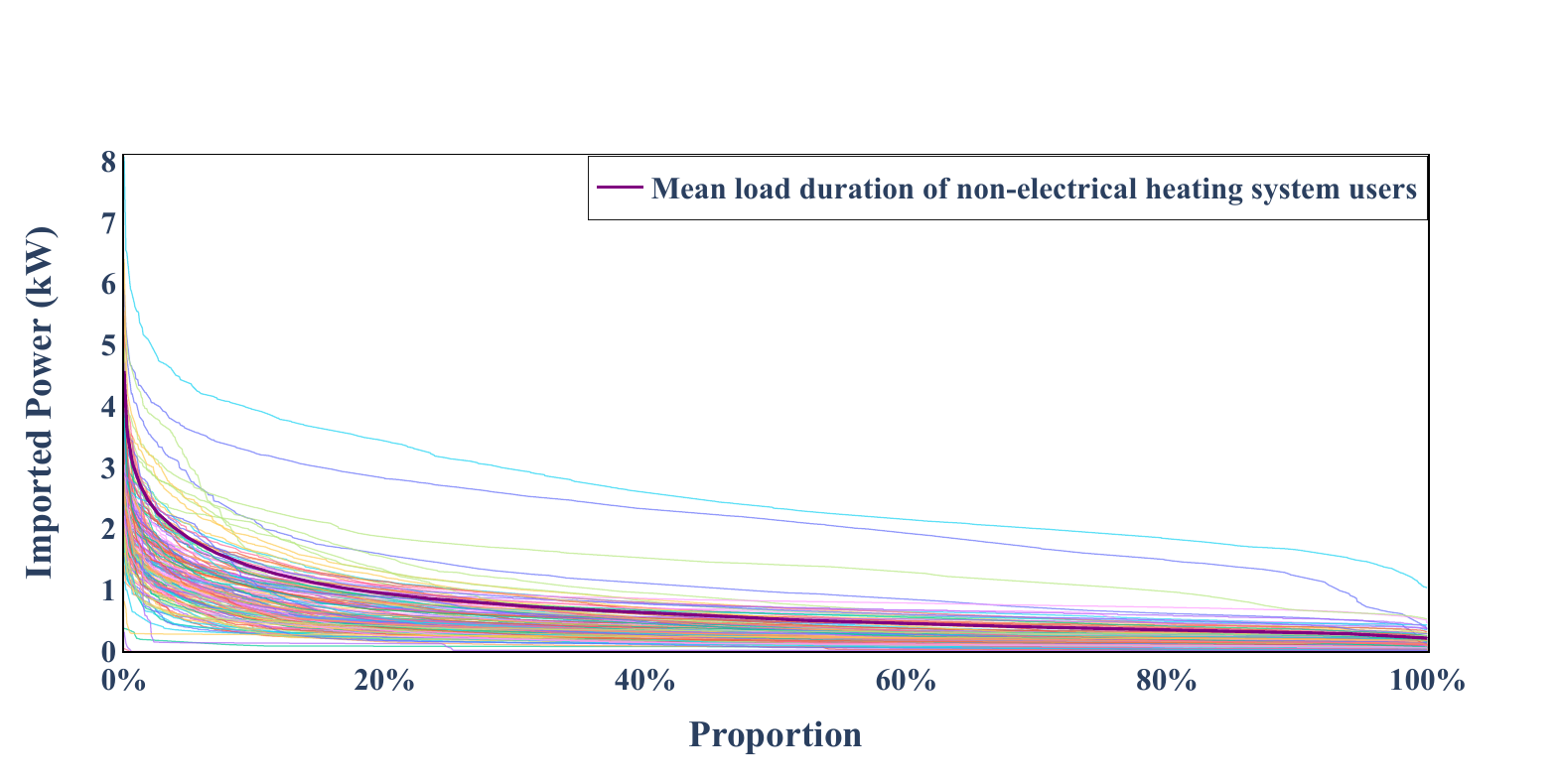}}
    \subfloat[Non-electric heating system users in summer\label{LD:summerne}]{\includegraphics[clip, trim=0.05cm 0.6cm 1.8cm 2.55cm,width=0.49\textwidth]{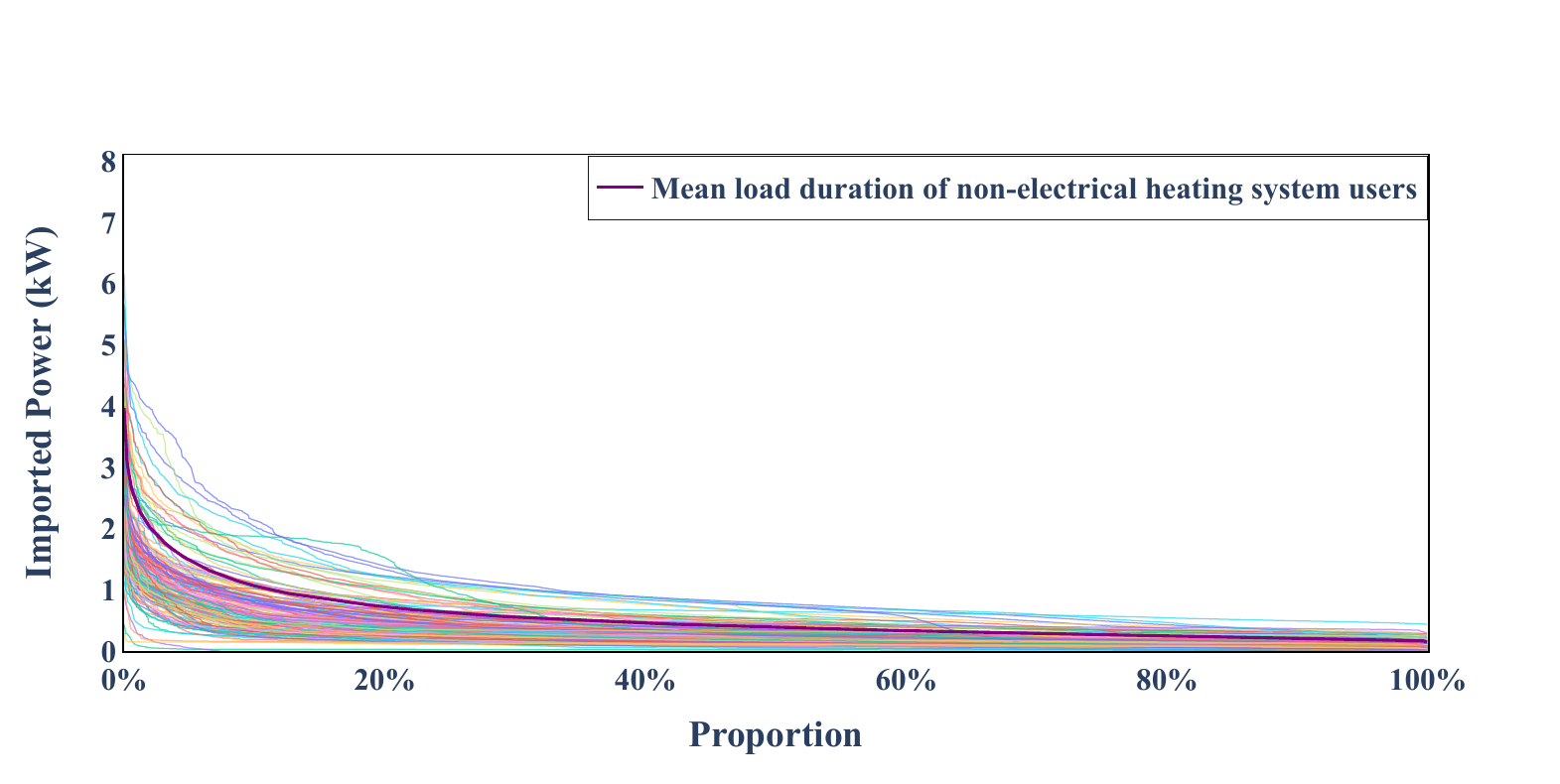}
    }
    \caption{Load duration plots for two seasons and two groups of users}\label{LD}
\end{figure*}

Since the dataset is imbalanced, stratified sampling is applied, i.e., we randomly selected 670 users with non-electric heating systems out of 893 and 120 users with electric heating systems out of 159 to build our training set. The remaining users are assigned to the testing set. For heating system classification, an intuitive method has been developed based on the fact that we expect electrical heating system users to import more energy from the grid during winter. To test this hypothesis, we first plot the load duration curves of each group of consumers during the summer and winter seasons separately, as shown in Figure \ref{LD}. 

\begin{figure}[H]
\centering
\includegraphics[clip, trim=2cm 0.1cm 2cm 2cm, width=3.7in]{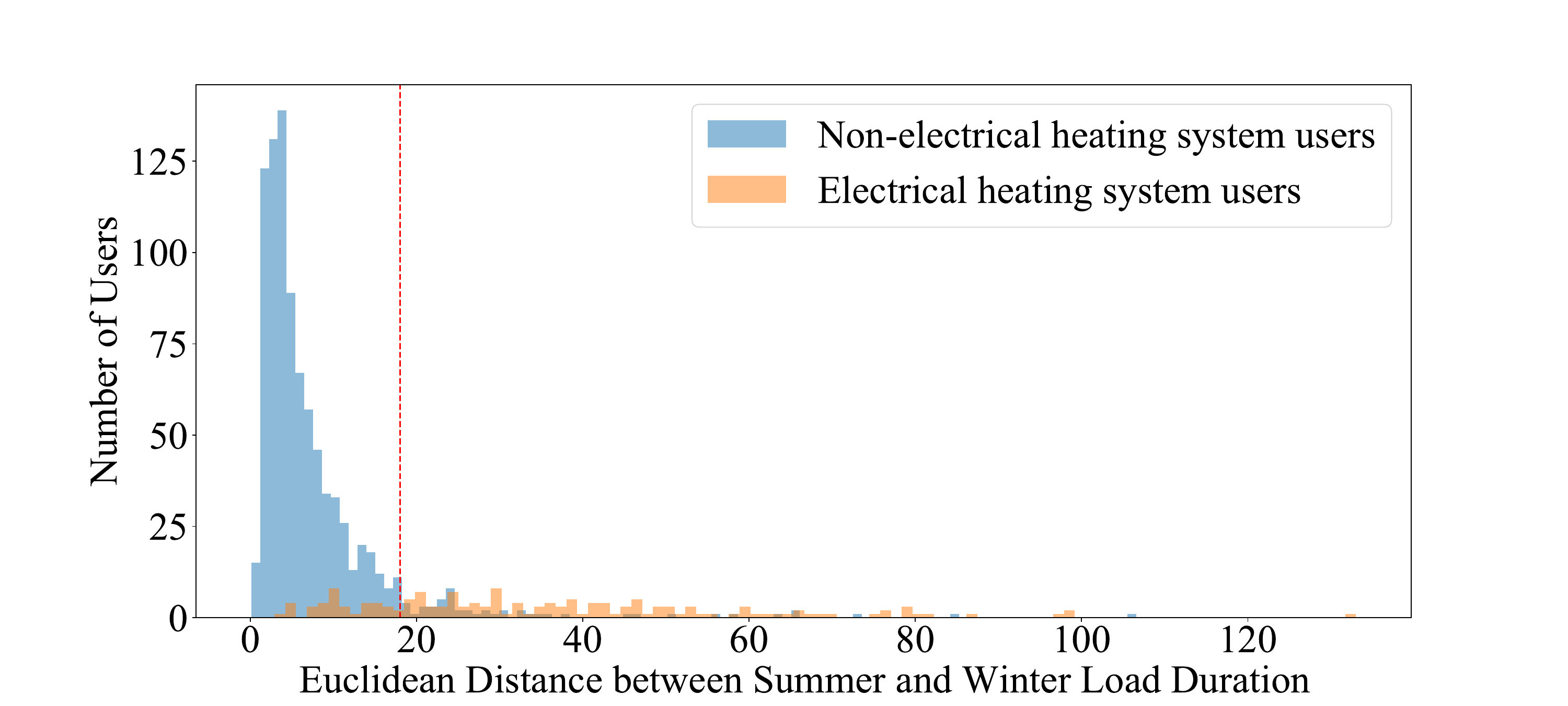}
\caption{A histogram of ED for load duration}
\label{KDE:LD}
\end{figure}

The higher imported energy can be observed for electrical heating system users compared to non-electrical heating users. The difference between summer load duration and winter load duration for each user is then measured by ED, and a threshold is set to identify the user's type, as shown in Figure \ref{KDE:LD}. This method is called the Load Duration (LD) method. 
For comparison, we use the same classification methods from the solar user's identification problem (except for the CZ that is replaced by the LD method). For obvious reasons, only winter data is used for the motif-related methods. The accuracy and computation time of the different methods are reported in Table~\ref{Comparision table: ES}. A comparison between Table~\ref{Comparision table: ES} and Table~\ref{Comparision table:PV} shows a lower F-score in the electric heating identification problem using IRMAC, which may be due to the low-resolution data as well as higher variations of electricity heating behaviours. Both tables show that the proposed IRMAC method is almost as accurate as the most accurate method, but it requires significantly less computation time on classification. Also, it proves that the proposed binary classification method is robust and could apply to other problems. 

\begin{table}[htb]
    \centering
    \caption{Performance of different methods in the electric heating system identification problem}
    \resizebox{1\textwidth}{!}{
    \begin{tabular}{*{10}{c}}
        \toprule
        \rowcolor{TBLHeader} & \multicolumn{4}{c}{Full ML} & \multicolumn{2}{c}{Intuitive} & \multicolumn{3}{c}{Motif-based} \\\cmidrule(l){2-5}\cmidrule(l){6-7}\cmidrule(l){8-10}
        \rowcolor{TBLHeader} & SVM & KNN & 1D-CNN & MLP & Average & LD & STOMP & DTW Motifs & \textbf{IRMAC} \\\midrule
         Accuracy & \textbf{95\%} & 93\% & \textbf{95\%} & \textbf{95\%} & 94\% & 92\% & 92\% & 90\% & 94.4\% \\ 
         \rowcolor{TBLRow}F-score & 0.79 & 0.72 & \textbf{0.8} & 0.767 & 0.73 & 0.73 & 0.642 & 0.622 & 0.749\\
         Time, s & 0.46 & 0.61 & 2.3 & 1.34 & \textbf{0.002} & 0.04 & \textbf{0.002} & \textbf{0.002} & \textbf{0.002} \\
         \rowcolor{TBLRow}Algorithmic transparency & \xmark & \cmark & \xmark & \xmark & \cmark & \cmark & \cmark & \cmark & \cmark \\
         \bottomrule
    \end{tabular}
    }
    \label{Comparision table: ES}
\end{table}

\section{Conclusion and Future Work}\label{future}
In this paper, we proposed a novel shape-based method to identify users with rooftop PV and electric heating systems from imported electricity data. This method is proven to be reliable and can be applied to other applications. Comparing the proposed IRMAC method with alternatives in Tables~\ref{Comparision table:PV} and \ref{Comparision table: ES} shows that it is accurate, fast, interpretable, and robust. While DNN-based methods, e.g., 1D-CNN and MLP, need large amounts of data to offer an unbiased solution, the IRMAC method only requires 24 hours of the users' RMs. 
Together with algorithmic transparency, these features are necessary for the wider acceptance of data-driven methods by practitioners. 
Furthermore, it is easy to maintain and upscale the IRMAC method in real-world applications since the motif of each user is extracted independently, and updating the data does not need re-computing RM in the entire dataset. Due to these advantages, it is acceptable to sacrifice a small amount of accuracy in exchange for higher interpretability, scalability, maintainability, and computation speed.

We also notice that there are some limitations to the current work. First, the amount of data that end users need to preserve for RM discovery might also raise security concerns on the users' end. Second, this method cannot identify the equipment that is scarcely used or with trivial influences on the electricity usage, as it relies on the patterns of the imported electricity. 

In future work, we plan to enhance the IRMAC by developing an adaptable motif technique to dynamically update motifs at the users' end, as well as exploring the minimum data required from the users' end to reduce the fears of security breaches. This way, consumer classification can be carried out on a daily basis, where faulty, shaded and under-performing PV systems can also be identified for the benefit of smart grid applications. Furthermore, we intend to develop an RM-based method for multi-label classification problems that have numerous applications in power systems, e.g., identifying users with electric vehicles, solar PV and stationary batteries.
\section{Acknowledgement}\label{acks}
This project is funded jointly by the University of Adelaide industry-PhD grant scheme and Watts, Denmark.

\bibliography{main}

\end{document}